\documentclass[10pt,twocolumn,letterpaper]{article}

 \usepackage{cvpr}              % To produce the CAMERA-READY version
\usepackage[dvipsnames]{xcolor}

\usepackage{multirow}
\usepackage{makecell}
\usepackage{booktabs}

\usepackage[linesnumbered,ruled,vlined]{algorithm2e}
\SetKwInput{KwInput}{Input}                % Set the Input
\SetKwInput{KwOutput}{Output}              % set the Output
\SetKw{KwBy}{by}

\newcommand*{\affaddr}[1]{#1} % No op here. Customize it for different styles.
\newcommand*{\affmark}[1][*]{\textsuperscript{#1}}

\definecolor{cvprblue}{rgb}{0.21,0.49,0.74}
\usepackage[pagebackref,breaklinks,colorlinks,citecolor=cvprblue]{hyperref}

\def\paperID{13555} % *** Enter the Paper ID here
\def\confName{CVPR}
\def\confYear{2024}

\title{DiffSal: Joint Audio and Video Learning for Diffusion Saliency Prediction}

\author{
	Junwen Xiong\affmark[1,2], 
	Peng Zhang\affmark[1,2]$^{*}$, 
	Tao You\affmark[1]\thanks{Corresponding author.},  
	Chuanyue Li\affmark[1], 
	Wei Huang\affmark[3], 
	Yufei Zha\affmark[1,2] \\
	\affaddr{\affmark[1]Northwestern Polytechnical University \\
	\affmark[2]Ningbo Institute of Northwestern Polytechnical University}\\
	\affaddr{\affmark[3]Nanchang University } \\
}

\begin{document}
\maketitle
\begin{abstract}
%Audio-visual saliency prediction can draw the support from diverse modality complement, but a further performance enhancement is still challenged by the customized architectures as well as task-specific loss functions. In recent studies, denoising diffusion model has shown to be more promisingly in unifying task frameworks owing to its intrinsic capability of generalization. Following this motivation, a novel \textbf{Diff}usion architecture for generalized audio-visual \textbf{Sal}iency prediction (DiffSal) is proposed in this work, which formulates the prediction problem as a conditional generative task of the saliency map by utilizing audio-visual inputs as the conditions. Based on the audio-visual feature extraction, an extra network `Saliency-UNet' is designed to perform multi-modal attention modulation for a progressive refinement of the ground-truth saliency map from the noisy map. Extensive experiments demonstrate that the proposed DiffSal is able to achieve excellent performance across six challenging audio-visual benchmarks, with an average relative improvement of 6.3\% over the previous state-of-the-art results by six metrics.

Audio-visual saliency prediction can draw support from diverse modality complements, but further performance enhancement is still challenged by customized architectures as well as task-specific loss functions. 
In recent studies, denoising diffusion models have shown more promising in unifying task frameworks owing to their inherent ability of generalization.
Following this motivation, a novel \textbf{Diff}usion architecture for generalized audio-visual \textbf{Sal}iency prediction (DiffSal) is proposed in this work, which formulates the prediction problem as a conditional generative task of the saliency map by utilizing input audio and video as the conditions. Based on the spatio-temporal audio-visual features, an extra network Saliency-UNet is designed to perform multi-modal attention modulation for progressive refinement of the ground-truth saliency map from the noisy map. Extensive experiments demonstrate that the proposed DiffSal can achieve excellent performance across six challenging audio-visual benchmarks, with an average relative improvement of 6.3\% over the previous state-of-the-art results by six metrics. The project url is \url{https://junwenxiong.github.io/DiffSal}.

\vspace{-15pt}

\end{abstract}     

\section{Introduction}
\label{sec:intro}

With the functionality of visual and auditory sensory systems, human beings can quickly focus on the most interesting areas during their daily activities. Such a comprehensive capability of visual attention in multi-modal scenarios has been explored by numerous researchers and referred to as an \textit{audio-visual saliency prediction} (AVSP) task. Based on the related techniques, many valuable practical applications have come into utility ranging from video summarization \cite{hu2017deep} and compression \cite{zhu2018spatiotemporal} to virtual reality \cite{7965634} and augmented reality \cite{sitzmann2018saliency}.

%Audio-visual saliency prediction has been extensively studied in computer vision, with the aim of estimating the most prominent area in dynamic audio and video scenes.

%The localization-based and the 3D convolution-based methods use pixel-wise  source localization and multi-scale feature interaction to predict the saliency object, respectively, and both of them methods use confusing combinations of optimization objective.

\begin{figure}
	\includegraphics[scale=0.4]{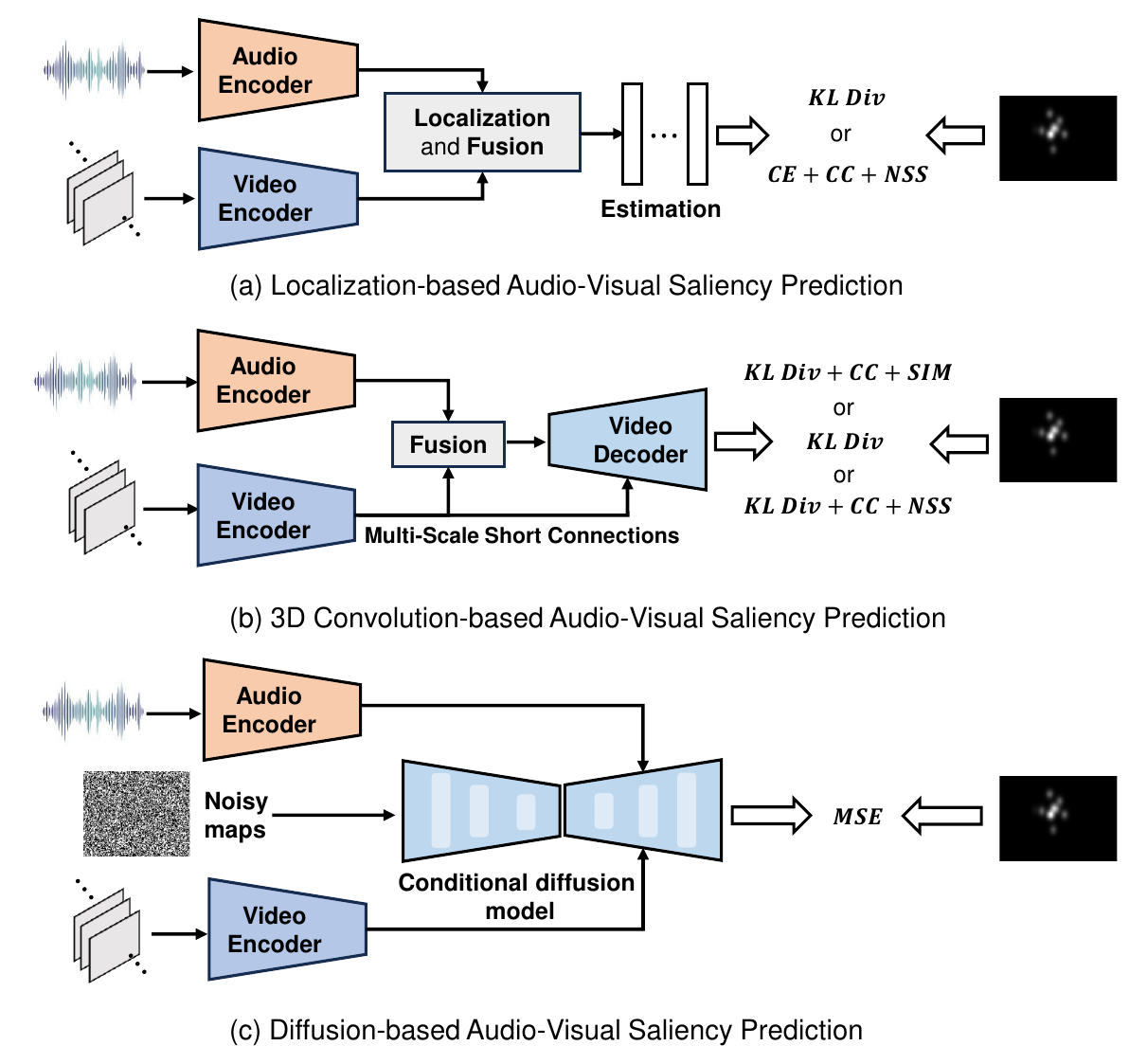}
	\vspace{-5pt}
	\caption{Comparison of conventional audio-visual saliency prediction paradigms and our proposed diffusion-based approach.
	Both the localization-based and 3D convolution-based methods use tailored network structures and sophisticated loss functions to predict saliency areas.
	Differently, our diffusion-based approach is a generalized audio-visual saliency prediction framework using simple \textit{MSE} objective function.}
	\vspace{-15pt}
	\label{fig-three-methods}
\end{figure}

% and sophisticated loss functions

Significant efforts have been dedicated to advancing studies by concentrating on elevating the quality of multimodal interaction and refining the generalizability of model structures in this domain. Among the prevalent AVSP approaches, as depicted in Figure \ref{fig-three-methods}(a), localization-based methods \cite{min2020multimodal, tsiami2020stavis, min2016fixation} have gained a lot of attention. These methods typically consider the sounding objects as saliency targets in the scene and transform the saliency prediction task into a spatial sound source localization problem. Even though the semantic interactions between audio and visual modalities have been considered in these methods, their focus on a generalized network structure is still limited and inevitably results in constrained performance.

In contrast, recent 3D convolution-based methods \cite{chang2021temporal, jain2021vinet, yao2021deep, xiong2023casp} exhibit superior performance in predicting audio-visual saliency maps, as illustrated in Figure \ref{fig-three-methods}(b). However, these methods require customized architectures with built-in inductive biases tailored for saliency prediction tasks. 
For instance, Jain \etal \cite{jain2021vinet} and Xiong \etal \cite{xiong2023casp} both embrace a 3D encoder-decoder structure akin to UNet, but integrate their empirical designs into the decoder. 
Moreover, both localization-based and 3D convolution-based methods employ sophisticated loss functions, contributing to a more intricate audio-visual saliency modeling paradigm.

Effective audio-visual interaction and the generalized saliency network are two essential factors for the seamless application of AVSP technology in the real world. Unfortunately, an in-depth exploitation of both challenges in the existing works is far from sufficient. 
Inspired by the strong generalization capabilities, diffusion models \cite{ho2020denoising, rombach2022high, ho2022classifier} can be employed as a \textit{unified} framework for generative tasks with class labels \cite{dhariwal2021diffusion}, text prompts \cite{gu2022vector}, images  \cite{chen2023diffusiondet}, and even sounds \cite{shen2023difftalk} as the conditions for modeling.
However, it remains an open question how to design a diffusion model that satisfies the effective audio-visual interaction and the generalized saliency network.

%However, it remains an open question as to how to utilize video and audio to guide the diffusion model to only highlight the most salient regions while suppressing the irrelevant ones.

%Moreover, to the best of our knowledge, there is no prior art that successfully adpots it to saliency prediction.

%To tackle these two challenges, we present a conditional \textbf{Diff}usion model for generalized audio-visual \textbf{Sal}iency prediction (DiffSal).

%, which aims to tackle these two challenges simultaneously,  as illustrated in Figure \ref{fig-three-methods}(c).

% Compared to GANs which have similar functionality, the diffusion model does not require the discriminator network, which is able to effectively avoid the common challenges such as unstable training and mode collapse in GANs.
%To tackle these issues, we present a conditional \textbf{Diff}usion model for generalized audio-visual \textbf{Sal}iency prediction (DiffSal), as shown in Figure \ref{fig-three-methods}(c).

%During training, the model is provided with the input video and audio cues as well as a degraded saliency map obtained from the ground-truth with varying degrees of injected noise.  

In this work, we present a conditional \textbf{Diff}usion model for generalized audio-visual \textbf{Sal}iency prediction (DiffSal), which aims to tackle these two challenges simultaneously,  as illustrated in Figure \ref{fig-three-methods}(c).
Our DiffSal utilizes the input video and audio as the conditions to reformulate the prediction problem as a conditional generative task of the saliency map.
During the training phase, the model is fed the video and audio cues as well as a degraded saliency map, which has been obtained from the ground-truth with varying degrees of injected noise.
By constructing a two-stream encoder to explore audio and video feature pairs with spatio-temporal coherence, the obtained similar pixel-wise multi-modal features can be utilized to guide the diffusion model generation process.
In addition, a novel network Saliency-UNet is employed to recover the original saliency maps from noisy inputs, which utilizes information from spatio-temporal audio and video features as the conditions.
To explore the latent semantic associations between audio and video features, an effective multi-modal interaction mechanism is proposed.
The entire DiffSal framework employs a \textit{simple} mean square error loss to predict ground-truth saliency maps from random noise.
During the inference phase, following the reversed diffusion process, DiffSal performs multi-step denoising to generate predictions based on randomly generated noisy saliency maps.

Benefiting from  such a diffusion-based framework, we demonstrate two distinct properties that appeal to the AVSP task.
(\textbf{i}) In contrast to existing methods with spatio-temporal visual branching \cite{tsiami2020stavis,xiong2023casp, jain2021vinet}, DiffSal enables spatio-temporal modeling of audio and video, and can be generalized to  audio-only, video-only, as well as audio-visual scenarios.
(\textbf{ii})  Thanks to the iterative denoising property of the diffusion model, DiffSal can iteratively reuse Saliency-UNet to improve performance without retraining.

To summarize, our main contributions are:
(1) We formulate the saliency prediction task as a conditional generative problem and propose a novel conditional diffusion saliency model, which is beneficial from the generalized network structure and effective audio-visual interaction.
(2) We demonstrate two properties of DiffSal that are effective on saliency prediction: the ability to be applied to either uni-modal or multi-modal scenarios, and to perform flexible iterative refinement without retraining.
(3) Extensive experiments have been conducted on six challenging audio-visual benchmarks and the results demonstrate that DiffSal achieves excellent performance, exhibiting an average relative improvement of 6.3\% over the previous state-of-the-art results across four metrics. 

\vspace{-5pt}
\section{Related Work}
\label{sec:related}

\subsection{Audio-Visual Saliency Prediction}
For audio-visual saliency prediction, different strategies for multi-modal correlation modeling have been proposed to estimate the saliency maps over consecutive frames. Early solutions \cite{min2016fixation, min2020multimodal} attempted to localize the moving-sounding target by canonical correlation analysis(CCA) to establish the cross-modal connections between the two modalities. With the advent of deep learning, Tsiami \etal \cite{tsiami2020stavis} continued the localization-based approach by extracting audio representation using SoundNet \cite{aytar2016soundnet}, and further performed spatial sound source localization through bilinear operations. Unfortunately, these methods exhibited sub-optimal performance only and thus led to the emergence of more effective 3D convolution-based approaches \cite{tavakoli2019dave, jain2021vinet, xiong2023casp} based on the encoder-decoder network frameworks. Jain \etal \cite{jain2021vinet} and Xiong \etal \cite{xiong2023casp} both embrace the UNet-style encoder-decoder structure by incorporating their empirical designs into the decoder. Moreover, Chang \etal \cite{chang2021temporal} employs a complex hierarchical feature pyramid network to aggregate deep semantic features. Considering that both the localization-based and 3D convolution-based methods use tailored network structures and sophisticated loss functions to predict saliency areas. In this study, by formulating the task as a conditional generation problem, a novel conditional diffusion model is proposed for generalized audio-visual saliency prediction.

%Early audio-visual saliency prediction methods attempted to establish the cross-modal connections between the two modalities by using CCA ,but an end-to-end deep learning scheme is still far from in-depth study. More recently, Tavakoli \etal \cite{tavakoli2019dave} propose to train two independent 3D ResNet for audio and visual modalities, and the outputs are directly concatenated as a late fusion strategy. With SoundNet \cite{aytar2016soundnet} for audio representation learning, Tsiami \etal\cite{tsiami2020stavis} perform a spatial sound source localization to obtain audio features, which are then fused with the visual feature maps by bilinear operation. In the same way, Jain \etal \cite{jain2021vinet} also employ bilinear fusion operation on the audio features of SoundNet and visual features to predict saliency maps. For these solutions based on a bilinear-based fusion scheme, the large number of learning parameters causes the model learning not easy to converge \cite{tenenbaum2000separating, yu2017multi}. In our work, an attention-based fusion is exploited to learn the cross-modal semantic interaction to overcome such a limitation.

%\subsection{Audio-Visual Learning}

\begin{figure*}[!htbp]
	\centering
	\includegraphics[width= \textwidth]{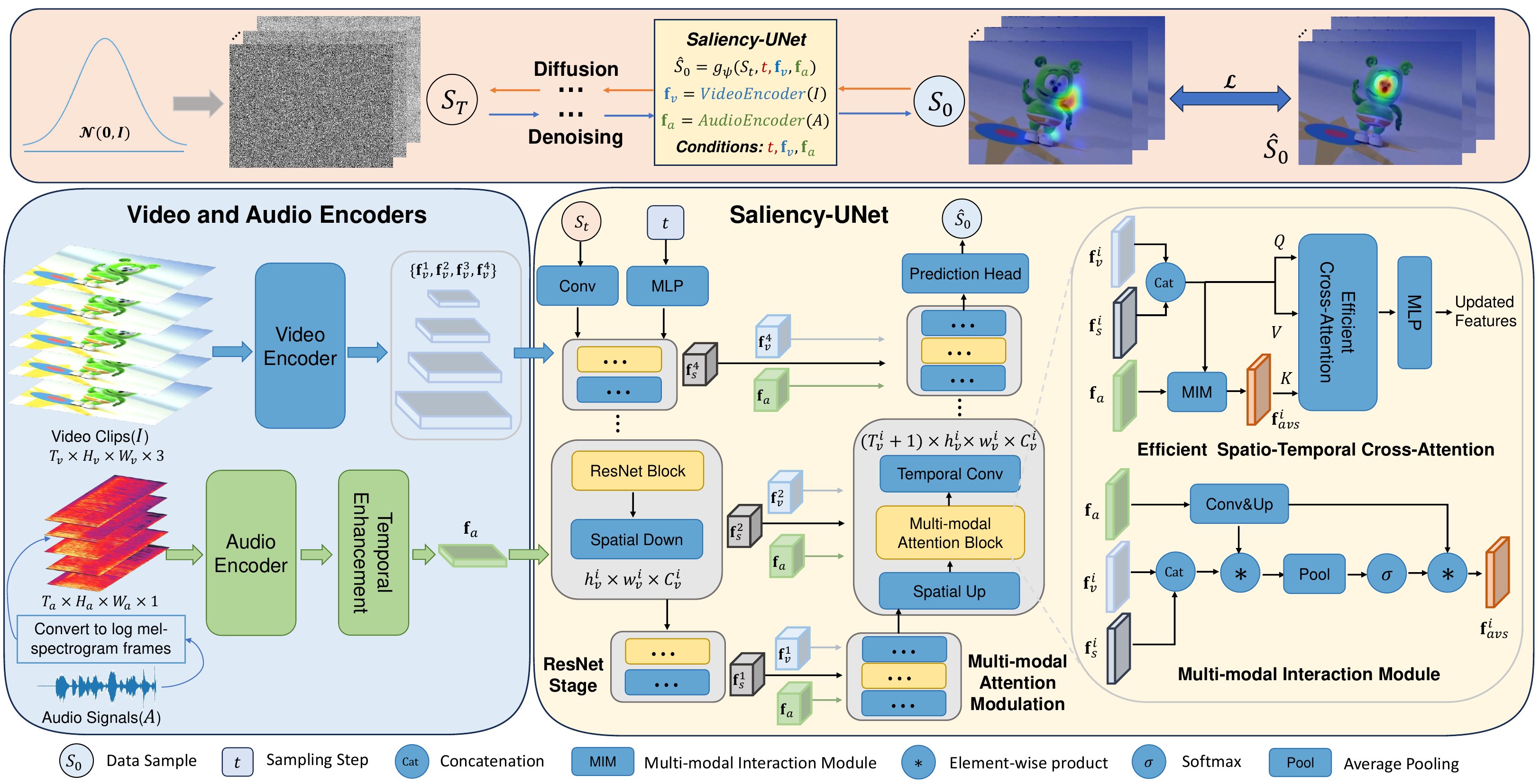}
	\vspace{-15pt}
	\caption{An overview of the proposed DiffSal framework. DiffSal first encodes  spatio-temporal video features $\textbf{f}_v$ and audio features $\textbf{f}_a$ by the Video and Audio Encoders, respectively.
		Then the Saliency-UNet takes audio features $\textbf{f}_a$  and video features $\textbf{f}_v$ as the conditions to guide the network in generating the saliency map $\hat{S}_0$ from the noisy map $S_t$.
	}
	\label{fig-model_structure}
	\vspace{-15pt}
\end{figure*}

\subsection{Diffusion Model}
Recently, diffusion models have gained significant traction in the field of deep learning. During diffusion modeling, the Markov process is employed to introduce noise into the training data followed by a training of deep neural networks to reverse it. Thanks to the high-quality generative results and strong generalization capabilities, diffusion models have achieved an impressive performance in generative tasks, such as image generation \cite{austin2021structured,benny2022dynamic,bond2022unleashing,choi2022perception,de2021diffusion}, image-to-image translation \cite{saharia2022palette,zhao2022egsde,wang2022pretraining,li2022vqbb,wolleb2022swiss}, video generation \cite{harvey2022flexible,yang2022diffusion,hoppe2022diffusion}, text-to-image synthesis \cite{gu2022vector,ramesh2022hierarchical,zhang2022fast}, and \etc. Beyond generative tasks, diffusion models have proven to be highly effective in various computer vision tasks. For instance, DiffSeg \cite{amit2021segdiff} proposes a diffusion model conditioned on an input image for image segmentation. Chen \etal \cite{chen2023diffusiondet} propose a model named DiffusionDet, which formulates object detection as a generative denoising process from noisy boxes to object boxes. Subsequently, the pipeline of this model is extended by Gu \etal \cite{gu2022diffusioninst}  by introducing noise filters during diffusion, as well as incorporating a mask branch for global mask reconstruction, which makes making DiffusionDet more applicable to instance segmentation tasks. To the best of our knowledge, there have been no previous successful attempts to apply diffusion models to saliency prediction, which inspires the proposed DiffSal in this work to explore the potential of diffusion models in the domain of audio-visual saliency prediction.

\vspace{-5pt}

\section{Preliminaries}
\label{sec:pre}

Diffusion models \cite{ho2020denoising} are likelihood-based models for points sampling from a given distribution by gradually denoising random Gaussian noise in $T$ steps. In the forward diffusion process, the increased noises are added to a sample point $x_0$ iteratively as $x_0 \to \cdots \to x_{T-1} \to x_{T}$, to obtain a completely noisy image $x_T$. Formally, the forward diffusion process is a Markovian noising process defined by a list of noise scales $\{ \bar{a}_t\}^T_{t=1}$ as:

\vspace{-5pt}
\begin{equation}
	\begin{aligned}
	&q(x_t|x_0) := \mathcal{N}(x_t | \sqrt{\bar{\alpha}_t}x_0, (1-\bar{\alpha}_t)\textit{\textbf{I}}), \\
	&x_t = \sqrt{\bar{\alpha}_t}x_0 + \sqrt{1-\bar{\alpha}_t}\epsilon, \epsilon \in \mathcal{N}(0, \textit{\textbf{I}}),
	\label{eq:diff_1}
	\end{aligned}
\end{equation}

\noindent where $ \bar{\alpha}_t := \prod_{s=1}^t \alpha_t = \prod_{s=1}^t (1-\beta_s)$ and $\beta_s$ denote the noise variance schedule \cite{ho2020denoising}, $\epsilon$ is the noise, $\mathcal{N}$ denotes normal distribution, $x_0$ is the original image, and $x_t$ is noisy image after $t$ steps of the diffusion process. 

The reverse diffusion process aims to learn the posterior distribution $p(x_{t-1} | x_0, x_t)$ for $x_{t-1}$ estimation given $x_t$. Typically, this can be done using a step-dependent neural network in multiple parameterized ways. Instead of directly predicting the noise $\epsilon$, we choose to parameterize the neural network $f_\theta(x_t, t)$ to predict $x_0$ as \cite{chen2023diffusiondet}. For model optimization, a mean squared error loss is employed to match $f_\theta(x_t, t)$ and $x_0$:

\vspace{-5pt}
\begin{equation}
	\mathcal{L} = \| f_\theta(x_t, t) - x_0 \|^2, t \in_R \{1,2, \dots, T\},
	\label{eq:diff_2}
\end{equation}

\noindent where the step $t$ is randomly selected at each training iteration.
From starting with a pure noise $x_t \in \mathcal{N}(0, \textit{\textbf{I}})$ during inference stage, the model can gradually reduce the noise according to the update rule \cite{song2020denoising} using the trained $f_\theta$ as below:

\vspace{-5pt}
\begin{equation}
	\begin{aligned}
	x_{t-1} = 	&\sqrt{\bar{\alpha}_{t-1}}f_\theta(x_t, t) +  \\
				&\sqrt{1-\bar{\alpha}_{t-1} - \sigma_t^2}\frac{x_t - \sqrt{\bar{\alpha}_t}f_\theta(x_t, t)}{\sqrt{1-\bar{\alpha}_t}} + \sigma_t \epsilon. \\
	\end{aligned}
	\label{eq:diff_3}
\end{equation}

Iteratively applying Eq. \ref{eq:diff_3}, a new sample $x_0$ can be generated from $f_\theta$ via a trajectory $x_T\to x_{T-1} \to \cdots \to x_0$. Specially, some improved sampling strategies skip such an operation in the trajectory to achieve better efficiency \cite{song2020denoising}.

To control the generation process, the conditional information can be modeled and incorporated in the diffusion model as an extra input $f_\theta(x_t, t, \mathbb{C})$. The class labels \cite{dhariwal2021diffusion}, text prompts \cite{gu2022vector}, and audio guidance \cite{ao2023gesturediffuclip} are the prevalent forms of conditional information documented in the literature.

\section{Method}

To tackle the challenges of effective audio-visual interaction and saliency network generalization, we formulate audio-visual saliency prediction as a conditional generation modeling of the saliency map, which treats the input video and audio as the conditions. Figure \ref{fig-model_structure} illustrates the overview of the proposed DiffSal, which contains parts of Video and Audio Encoders as well as Saliency-UNet.
The former is used to extract multi-scale spatio-temporal video features and audio features  from image sequences and corresponding audio signals. By conditioning on these semantic video and audio features, the latter  performs multi-modal attention modulation to progressively refine the ground-truth saliency map from the noisy map. Each part of DiffSal is elaborated on below.

%Given the input video data $V \in  \mathbb{R}^{   T_v \times H_v  \times W_v \times  3 }$ and audio data $A \in \mathbb{R}^{   T_a \times H_a  \times W_a \times  1 }$,  we model audio-visual global correlations as conditions and approximate the data distribution of the ground truth saliency maps $H_0 \in \{0,1\}^{H_v \times W_v \times 1}$ based on these conditions .  The proposed DiffSal is composed of: a two-stream network to obtain visual saliency and auditory saliency feature, an audio-visual interaction module to integrate the visual and auditory conspicuity maps, a consistency-aware predictive coding module to reason the coherent spatio-temporal visual feature with audio feature, and a saliency decoder to estimate saliency map with multi-scale audio-visual features.

\subsection{Video and Audio Encoders}
\label{sec:av_encoders}

\textbf{Video Encoder}. Let $I = [I_1, \cdots, I_j, \cdots, I_{T_v}], I_j \in  \mathbb{R}^{H_v  \times W_v \times  3 }$ denotes an RGB video clip of length $T_v$.  This serves as the input to a video backbone network, which produces spatio-temporal feature maps. The backbone consists of 4 encoder stages and outputs 4 hierarchical video feature maps, illustrated in Figure \ref{fig-model_structure}. The generated feature maps are denoted as $\{\textbf{f}_{v}^{i} \}_{i=1}^{N}  \in \mathbb{R}^{ T^i_v \times h^i_v  \times w^i_v \times C_v^i} $, where $(h_v^i, w_v^i) = (H_v, W_v)/2^{i+1}, N=4$. In practical implementation, we employ the off-the-shelf MViTv2 \cite{li2022mvitv2} as a video encoder to encode the spatial and temporal information of image sequences. More generally, MViTv2 can also be replaced with other general-purpose encoders, \eg, S3D\cite{xie2018rethinking}, Video Swin Transformer\cite{liu2022video}.

% by first transforming the raw audio into a log-mel spectrogram via STFT, and then partitioning it into  $T_a$ slices of dimension $H_a \times W_a \times 1$  with the hop-window size of 11 ms.
\noindent \textbf{Audio Encoder}. To temporally synchronize the audio features with the video features in a better way, initially, we transform the raw audio into a log-mel spectrogram through Short-Time Fourier Transform (STFT). Then, the spectrogram is partitioned into $T_a$ slices of dimension $H_a \times W_a \times 1$ with a hop-window size of 11 ms.
To extract per-frame audio feature $\bar{\textbf{f}}_{a, i}$ where $i \in \{1, \cdots, T_a\}$, a pre-trained 2D fully convolutional VGGish network \cite{hershey2017cnn} is performed on AudioSet \cite{gemmeke2017audio}, resulting in a feature map of size $\mathbb{R}^{h_a \times w_a \times C_a}$. To improve the inter-frame consistency, we further introduce a temporal enhancement module consisting of a patch embedding layer as well as a transformer layer. Then,
the audio features are rearranged in the spatio-temporal dimension and fed into the patch embedding layer to obtain $\bar{\textbf{f}}_{a} \in \mathbb{R}^{T_a \times h_a \times w_a \times C_a}$. For the retaining of temporal position information, a learnable positional embedding $\mathrm{\textbf{e}}^{pos}$ is incorporated along the temporal dimension:
\vspace{-5pt}
\begin{equation}
	\begin{aligned}
		%		& \bar{f}_a = \mathrm{VGGish(A)}, \\
		%		& \bar{f}_a = \mathrm{PatchEmbed}(\bar{f}_a) + \mathrm{E_{POS}}, \\
		%		& f_a = \mathrm{TransEncoder}(\bar{f}_a), \\
		%		& \bar{f}_a = \mathrm{PatchEmbed}(\bar{f}_a)  \\
		&  \bar{\textbf{f}}_a = [\bar{\textbf{f}}_{a,0} + \mathrm{\textbf{e}}^{pos}_0, \cdots, \bar{\textbf{f}}_{a,{T_a}} + \mathrm{\textbf{e}}^{pos}_{T_a}],
		%		& f_a = \mathrm{TransEncoder}(\bar{f}_a), \\
	\end{aligned}
\end{equation}

\noindent where $[\cdot, \cdot]$ represents concatenation operation. The processed feature is finally fed into the Multi-head Self Attention (MSA), the layer normalization (LN) \cite{ba2016layer} and the MLP layer to produce the spatio-temporal audio features $\textbf{f}_a \in \mathbb{R}^{T_a \times h_a \times w_a \times C_a}$:

\vspace{-5pt}
\begin{equation}
	\begin{aligned}
		%		& \bar{f}_a = \mathrm{VGGish(A)}, \\
		%		& \bar{f}_a = \mathrm{PatchEmbed}(\bar{f}_a) + \mathrm{E_{POS}}, \\
		%		& f_a = \mathrm{TransEncoder}(\bar{f}_a), \\
		%		& \bar{f}_a = \mathrm{PatchEmbed}(\bar{f}_a)  \\
%		& \bar{f}_a = [\bar{f}_{a,0} + \mathrm{\textbf{e}}^{pos}_0, \cdots, \bar{f}_{a,{T_a}} + \mathrm{\textbf{e}}^{pos}_{T_a}] \\
		& \bar{\textbf{f}}_a = \mathtt{MSA(LN(}\bar{\textbf{f}}_a)) + \bar{\textbf{f}}_a, \\
		& \textbf{f}_a = \mathtt{MLP(LN(}\bar{\textbf{f}}_a)) + \bar{\textbf{f}}_a.
%		& f_a = \mathrm{TransEncoder}(\bar{f}_a),
	\end{aligned}
\end{equation}

\subsection{Saliency-UNet}
\label{sec:sal_decoder}

To learn the underlying distribution of saliency maps, we design a novel conditional denoising network $g_\psi$ with multi-modal attention modulation, named Saliency-UNet.
This network is designed to leverage both audio features $\textbf{f}_a$ and video features  $\{\textbf{f}_{v}^{i} \}_{i=1}^{N} $ as the conditions, guiding the network in generating the saliency map $\hat{S}_0$ from the noisy map $S_t$:

%Building upon the denoising process outlined in Section \ref{sec:pre}, 

% which can take audio feature $f_a$  and video feature $f_v$ as conditions to guide the network in generating the saliency map $x_0$ from the noisy map $x_t$.
\vspace{-5pt}
\begin{equation}
	\begin{aligned}
		& \hat{S}_0 = g_\psi(S_t, t, \textbf{f}_a, \textbf{f}_v),
	\end{aligned}
\end{equation}

\noindent where	$S_t = \sqrt{\bar{\alpha}_t}S_0 + \sqrt{1-\bar{\alpha}_t}\epsilon$, noise $\epsilon$ is from a Gaussian distribution, and $t \in_R \{1,2, \dots, T\}$ is a random diffusion step. 

Our Saliency-UNet can be functionally divided into two parts: feature encoding and feature decoding, as shown in Figure \ref{fig-model_structure}. 
The first part encodes multi-scale noise feature maps $\{\textbf{f}_{s}^{i}\}_{i=1}^{N} \in \mathbb{R}^{h_v^i \times w_v^i \times C_v^i}$ from the noisy map $S_t$ using multiple ResNet stages. The second part utilizes our designed multi-modal attention modulation (MAM) across multiple scales for the interaction of noise features, audio features, and video features. 
The MAM stage comprises an upsampling layer, a multi-modal attention block, and a 3D temporal convolution. This stage not only computes the global spatio-temporal correlation between multi-modal features but also progressively enhances the spatial resolution of the feature maps. 
At last, a prediction head is employed to produce the predicted saliency map $\hat{S}_0$.
The entire network incorporates 4 layers of the ResNet stage for feature encoding and 4 layers of the MAM stage for feature decoding.

%The MAM structure not only enables interactions between multi-modal features, but also computes global spatio-temporal correlations between audio and visual features
%Our Saliency-UNet is structurally similar to the original diffusion model \cite{ho2020denoising}, as shown in Figure \ref{fig-model_structure}, the difference being that our network replaces the standard ResNet stages in the decoder part with the designed multi-modal attention modulation stages. This choice of structure facilitates the computation of global spatio-temporal correlations between audio and visual features, while also enabling interaction between multi-modal features. The decoder part of the  Saliency-UNet network employs four multi-modal attention modulation stages, each of which consists of an upsampling operation, a multi-modal attention block, and a temporal convolution operation, respectively.  
% the Saliency-UNet encoder extracts multi-scale noisy features $\{f_{x}^{i}\} (i=1, \dots, 4)$ from the noisy map $x_t$. They are then fed into the Saliency-UNet decoder along with the acquired multi-scale visual features $\{f_{v}^{i} \}(i=1, \dots, 4)$ and audio features $f_a$.  During the decoding process, we gradually incorporate interacted multimodal features and progressively reconstruct the resolution of the features until generating the saliency map.
 
For more robust multi-modal feature generation, two techniques in MAM are introduced: efficient spatio-temporal cross-attention  and multi-modal interaction module.

\noindent \textbf{Efficient  Spatio-Temporal Cross-Attention}. Given video features $\textbf{f}_{v}^{i} \in \mathbb{R}^{T_v^i \times h_v^i \times w_v^i \times C_v^i }$, audio features $\textbf{f}_a \in \mathbb{R}^{T_a \times h_a \times w_a \times C_a}$ and noise features $\textbf{f}_{s}^{i} \in \mathbb{R}^{h_v^i \times w_v^i \times C_v^i}$,  the video features and noise features are concatenated as a spatio-temporal noise feature map $[\textbf{f}_{v}^i, \textbf{f}_{s}^i]\in \mathbb{R}^{(T_v^i +1) \times h_v^i \times w_v^i \times C_v^i }$. The processed feature map is then fed into the multi-modal interaction module along with the audio features to obtain $\textbf{f}_{avs} \in \mathbb{R}^{(T_v^i +1) \times h_v^i \times w_v^i \times C_v^i }$. 
To introduce audio information in the saliency prediction, we design an efficient spatio-temporal cross-attention (ECA), which not only effectively reduces the computational overhead of the standard cross-attention \cite{vaswani2017attention}, but also possesses spatio-temporal interactions between features.

For all processed features, they are first converted to 2D feature sequences in the spatio-temporal domain and obtained ${Q} = {V} = [\textbf{f}_{v}^i, \textbf{f}_{s}^i] \in \mathbb{R}^{(T_v^i +1) h_v^i  w_v^i \times C_v^i}, {K} = \textbf{f}_{avs}^i \in \mathbb{R}^{(T_v^i +1) h_v^i  w_v^i \times C_v^i} $ . We find that directly  following the standard cross-attention would take a prohibitively large amount of memory due to the high spatio-temporal resolution of the feature map. Therefore, a spatio-temporal compression  technique is employed that can effectively reduce the computational overhead without compromising performance, as:

%After acquiring all the processed features, we find that directly  following the standard cross-attention \cite{vaswani2017attention} would take a prohibitively large amount of memory due to the high spatio-temporal resolution of the feature map. Inspired by spatial-reduction attention in PVT \cite{wang2021pyramid}, we directly apply a spatio-temporal reduction operation  to key ${K}$ and value ${V}$, which drastically reduces the computation and memory overhead, as follows: 

\vspace{-5pt}
\begin{equation}
	\begin{aligned}
%		{Q} &= {V} = [\textbf{f}_{v}^i, \textbf{f}_{s}^i], {K} = \textbf{f}_{avs}^i, \\
		[\textbf{f}_{v}^i, \textbf{f}_{s}^i] &= \mathtt{ECA}( Q W_Q, \mathtt{STC}(K)W_K, \mathtt{STC}(V)W_V),
	\end{aligned}
\end{equation}

\noindent where $W_Q, W_K, W_V \in \mathbb{R}^{C_i \times C_i} $ are parameters of linear projections, $\mathtt{STC(\cdot)}$ is the spatio-temporal compression, which is defined as:

\vspace{-5pt}
\begin{equation}
	\begin{aligned}
		& \mathtt{STC}(x) = \mathtt{LN(Conv3d}(x)),
	\end{aligned}
\end{equation}

\noindent Here, the dimension of features is reduced by controlling the kernel size as well as the stride size of 3D convolution.

\noindent \textbf{Multi-modal Interaction Module}. To take full advantage of different modal features, we model the interaction between audio features $\textbf{f}_a$, video features $\textbf{f}_v^{i}$ and noise features $\textbf{f}_{s}^i$ at each scale $i$ to obtain a robust multi-modal feature representation. A straightforward approach would be to directly concatenate and aggregate $\textbf{f}_a$, $\textbf{f}_v^{i}$ and $\textbf{f}_{s}^i$, but this fusion method does not acquire global correlations between various modalities.
Therefore, an effective multi-modal interaction strategy is proposed to capture crucial audio-visual activity changes in the spatio-temporal domain.
In specific, this process starts with convolution and upsampling on the audio features, which is to construct spatially size-matched feature triples $(\widetilde{\textbf{f}}_a, \textbf{f}_v^i, \textbf{f}_s^i)$.
Subsequently, the video features and noise features are concatenated to obtain spatio-temporal noise features $[\textbf{f}_{v}^i, \textbf{f}_{s}^i]$. This processed feature undergoes an element-wise product operation with the audio features, resulting in $\bar{\textbf{f}}_{avs}^i$. Following this, we perform average activation along the temporal dimension over $\bar{\textbf{f}}_{avs}^i$ to pool global temporal information into a temporal descriptor. For the indication of critical motion regions, a softmax function is applied to obtain a mask by highlighting the segments of the corresponding spatio-temporal audio features, which exhibit key audio-visual activity changes:

%The video feature and noisy feature are then concatenated together to obtain spatio-temporal noisy features $[f_{v}^i; f_{x}^i]$. This processed feature is subjected to an element-wise product operation with the audio feature to yield $\bar{f}_{avx}^i$. Subsequently, we take the average activation along the temporal dimension over $\bar{f}_{avx}^i$ to pool the global temporal information into a temporal descriptor, followed by a softmax function to obtain a mask that indicates the critical motion region. Finally, the mask is utilized to highlight segments of the corresponding spatio-temporal noise features that have key audio-visual activity changes.
\vspace{-5pt}
\begin{equation}
	\begin{aligned}
		 \widetilde{\textbf{f}}_a &= \mathtt{Conv}(\mathtt{UpSample}(\textbf{f}_a)),\\
		\textbf{f}_{avs}^i &= \mathtt{softmax}(\mathtt{Pool}([\textbf{f}_{v}^i, \textbf{f}_{s}^i] \ast \widetilde{\textbf{f}}_a)) \ast \widetilde{\textbf{f}}_a.
	\end{aligned}
\end{equation}

\noindent where $\mathtt{Conv}(\cdot)$,  $\ast$, $\mathtt{softmax}$ and $\mathtt{Pool}$ denote the operations of convolution, element-wise product, softmax and average pooling, respectively.

%\noindent \textbf{Saliency Map Generation}. 
%Finally, we employ a prediction head to yield the predicted saliency map $\hat{Y}_t$. 
%
%Ultimately, we integrate feature representations across all scales $\{\mathcal{F}_1,  \mathcal{F}_2, \mathcal{F}_3, \mathcal{F}_4 \}$ via element-wise addition, and employ a prediction head to yield the predicted saliency map $\hat{Y}_t$. By inductively modeling multi-scale feature interactions, our Saliency-UNet is able to reason about the fine-grained relationships between key saliency regions and global contexts, thereby producing more tailored representations that attend to joint audio-visual activity.
%

%We initiate a diffusion process that introduces corruption to ground-truth saliency maps, creating noisy maps. Subsequently, we train the Saliency-UNet for saliency map denoising, aiming to reverse this process. The overall training procedure of our DiffSal is outlined in Algorithm \ref{alg:training} in the Appendix. In detail, we sample Gaussian noises following $\alpha_t$ in Eq. \ref{eq:diff_1} and add them to ground-truth saliency maps, resulting in noisy samples. The parameter $\alpha_t$ at each sampling step $t$ is pre-defined by a monotonically decreasing cosine scheme, as employed in \cite{ho2020denoising}. The standard mean square error serves as the optimization function to supervise the model training:

\subsection{Overall Training and Inference Algorithms}
\label{sec:training}

\textbf{Training}. A diffusion process is initiated to create noisy maps by introducing corruption to ground-truth saliency maps. To reverse this process, the Saliency-UNet is trained for saliency map denoising. The overall training procedure of DiffSal is outlined in Algorithm \ref{alg:training} in the Appendix. In detail, Gaussian noises are sampled following $\alpha_t$ in Eq. \ref{eq:diff_1} and added to the ground-truth saliency maps, resulting in noisy samples. At each sampling step $t$, the parameter $\alpha_t$ is pre-defined by a monotonically decreasing cosine scheme, as employed in \cite{ho2020denoising}. The standard mean square error serves as the optimization function to supervise the model training:
\vspace{-5pt}
\begin{equation}
	\begin{aligned}
		& \mathcal{L} =  \| S_0 - g_\psi(S_t, t, \textbf{f}_a, \textbf{f}_v) \|^2.
	\end{aligned}
\end{equation}

\noindent where $S_0$ and $g_\psi(S_t, t, \textbf{f}_a, \textbf{f}_v)$ denote the ground-truth and predicted saliency maps, respectively.

\noindent \textbf{Inference}. The proposed DiffSal engages in denoising noisy saliency maps sampled from a Gaussian distribution, and progressively refines the corresponding predictions across multiple sampling steps. In each sampling step, the Saliency-UNet processes random noisy saliency maps or the predicted saliency maps from the previous sampling step as input and generates the estimated saliency maps for the current step. For the next step, DDIM \cite{song2020denoising} is applied to update the saliency maps. The detailed inference procedure is outlined in Algorithm \ref{alg:infer}.

\section{Experiments}
Experiments are conducted on six audio-visual datasets. The following subsections introduce the implementation details and evaluation metrics. The experimental results are represented with analysis through ablation studies and comparison with state-of-the-art works.

\subsection{Setup}
\noindent \textbf{Audio-Visual Datasets}: Six audio-visual datasets in saliency prediction have been employed for the evaluation, which are: AVAD \cite{min2016fixation}, Coutrot1 \cite{coutrot2014saliency}, Coutrot2 \cite{coutrot2016multimodal}, DIEM \cite{mital2011clustering}, ETMD \cite{koutras2015perceptually}, and SumMe \cite{gygli2014creating}. The significant characteristics of these datasets are elaborated below.
(i) The AVAD dataset comprises 45 video clips with durations ranging from 5 to 10 seconds. These clips cover various audio-visual activities, such as playing the piano, playing basketball, conducting interviews, \etc. This dataset contains eye-tracking data from 16 participants.
(ii) The Coutrot1 and Coutrot2 datasets are derived from the Coutrot dataset. Coutrot1 consists of 60 video clips covering four visual categories: one moving object, several moving objects, landscapes, and faces. The corresponding eye-tracking data are obtained from 72 participants. Coutrot2 includes 15 video clips recording four individuals having a meeting, with eye-tracking data from 40 participants.
(iii) The DIEM dataset contains 84 video clips, including game trailers, music videos, advertisements, \etc., captured from 42 participants. Notably, the audio and visual tracks in these videos do not naturally correspond.
(iv) The ETMD dataset comprises 12 video clips extracted from various Hollywood movies, with eye-tracking data annotated by 10 different persons.
(v) The SumMe dataset consists of 25 video clips covering diverse topics, such as playing ball, cooking, travelling, \etc.

%There are six audio-visual datasets in saliency prediction, AVAD \cite{min2016fixation}, Coutrot1 \cite{coutrot2014saliency}, Coutrot2 \cite{coutrot2016multimodal}, DIEM \cite{mital2011clustering}, ETMD \cite{koutras2015perceptually}, and SumMe \cite{gygli2014creating}, used for our evaluation and comparison. In detail:
%(i) The AVAD dataset comprises 45 video clips with durations ranging from 5 to 10 seconds. These clips cover various audio-visual activities, such as playing the piano, playing basketball, conducting interviews, \etc. The dataset includes eye-tracking data from 16 participants.
%(ii) The Coutrot1 and Coutrot2 datasets are derived from the Coutrot dataset. Coutrot1 consists of 60 video clips covering four visual categories: one moving object, several moving objects, landscapes, and faces. The corresponding eye-tracking data are obtained from 72 participants. Coutrot2 includes 15 video clips recording four individuals having a meeting, with eye-tracking data from 40 participants.
%(iii) The DIEM dataset contains 84 video clips, including game trailers, music videos, advertisements, \etc., captured from 42 participants. Notably, the audio and visual tracks in these videos do not naturally correspond.
%(iv) The ETMD dataset comprises 12 video clips extracted from various Hollywood movies, with eye-tracking data annotated by 10 different people.
%(v) The SumMe dataset consists of 25 video clips covering diverse topics, such as playing ball, cooking, traveling, \etc. 

\begin{table}[t]
	\centering
	\renewcommand\arraystretch{1.2}
	\caption{Ablation of different components in DiffSal.}
	\vspace{-5pt}
	\resizebox{0.85 \linewidth}{!}{
		\begin{tabular}{l ll llll}
			\toprule
			%			\multirow{2}{*}{Method} & \multirow{2}{*}{ECAM} & \multirow{2}{*}{MIM} & \multicolumn{2}{c}{\textbf{AVAD}}  & \multicolumn{2}{c}{\textbf{\textrm{ETMD}}}  \\
			\multirow{2}{*}{Method} & \multicolumn{2}{c}{Components} & \multicolumn{2}{c}{\textbf{AVAD}}  & \multicolumn{2}{c}{\textbf{\textrm{ETMD}}} \\
			\cmidrule(r){2-3} \cmidrule(r){4-5}  \cmidrule(r){6-7}
			& 			ECA& 		MIM& 	CC $\uparrow$    & SIM $\uparrow$ & CC $\uparrow$   & SIM $\uparrow$  \\
			\midrule
			baseline      &             	&      			 & 0.701 		 & 0.547          & 0.632           & 0.498     \\
			& \checkmark  &         	    & 0.716			 & 0.556          & 0.644           & 0.503    \\
			&   		 &   \checkmark 	& 0.714			 & 0.551          & 0.638           & 0.502   \\
			& \checkmark &  \checkmark   	& \textbf{0.738} & \textbf{0.571} & \textbf{0.652}  & \textbf{0.506}    \\
			\bottomrule
	\end{tabular}}
	\vspace{-10pt}
	\label{table-ablation_1}
\end{table}

%To facilitate implementation, we use pre-trained  MViTv2  model \cite{li2022mvitv2} on Kinetics \cite{carreira2017quo} and pre-trained VGGish \cite{hershey2017cnn} on AudioSet \cite{gemmeke2017audio}. 

%\noindent \textbf{Implementation Details}: To facilitate implementation, a pre-trained MViTv2 \cite{li2022mvitv2} model on Kinetics \cite{carreira2017quo} and a pre-trained VGGish \cite{hershey2017cnn} on AudioSet \cite{gemmeke2017audio} are used.
%The input samples of the network consist of 16-frame video clips of size $224 \times 384 \times 3$ with the corresponding audio, which is transformed into $9$ slices  of $112 \times 192$ log-Mel spectrograms. 
%For the spatio-temporal compression in efficient spatio-temporal cross-attention, 
%the kernel size and stride size of the 3D convolution in the $i$th MAM stage are set to $2^i$ and $2^i$, respectively.
%For a fair comparison, following \cite{xiong2023casp}, the video branch of DiffSal is pre-trained using the DHF1k dataset \cite{wang2018revisiting}, and then the entire model is fine-tuned on these audio-visual datasets using this pre-trained weight.
% The training process chooses Adam as the optimizer with the started learning rate of 1e-4. The computation platform is configured by four NVIDIA GeForce RTX 4090 GPUs in a distributed fashion, using Pytorch. 
%The total sampling step $T$ is defined as $1000$. All training process is terminated within 5 epochs. The batch size for the entire experiment is set to 20. During inference, the iterative denoising step is set to 4.

\noindent \textbf{Implementation Details}: To facilitate implementation, a pre-trained MViTv2 \cite{li2022mvitv2} model on Kinetics \cite{carreira2017quo} and a pre-trained VGGish \cite{hershey2017cnn} on AudioSet \cite{gemmeke2017audio} are adopted. The input samples of the network consist of 16-frame video clips of size $224 \times 384 \times 3$ with the corresponding audio, which is transformed into $9$ slices  of $112 \times 192$ log-Mel spectrograms. For the spatio-temporal compression in efficient spatio-temporal cross-attention, the kernel size and stride size of the 3D convolution in the $i$th MAM stage are set to $2^i$ and $2^i$, respectively. For a fair comparison, the video branch of DiffSal is pre-trained using the DHF1k dataset \cite{wang2018revisiting} following \cite{xiong2023casp}, and the entire model is fine-tuned on these audio-visual datasets using this pre-trained weight.

The training process chooses Adam as the optimizer with the started learning rate of $1e-4$. The computation platform is configured by four NVIDIA GeForce RTX 4090 GPUs in a distributed fashion, using Pytorch. The total sampling step $T$ is defined as $1000$ and the entire training is terminated within 5 epochs. The batch size is set to 20 across all experiments. During inference, the iterative denoising step is set to 4.

\noindent \textbf{Evaluation Metrics}: Following previous works, four widely-used evaluation metrics are adopted \cite{bylinskii2018different}: CC, NSS, AUC-Judd (AUC-J), and SIM.  The same evaluation codes are used as in previous works \cite{tsiami2020stavis,xiong2023casp}.

\subsection{Ablation Studies}
Extensive ablation studies are performed to validate the design choices in the method. The AVAD and ETMD datasets are selected for ablation studies, following the approach in \cite{xiong2023casp}.

\begin{table}[t]
	\centering
	\renewcommand\arraystretch{1.2}
	\caption{Ablation of video and audio modalities.}
	\vspace{-5pt}
	\resizebox{0.7 \linewidth}{!}{
		\begin{tabular}{l llll}
			\toprule
			\multirow{2}{*}{Model} &  \multicolumn{2}{c}{\textbf{AVAD}}  & \multicolumn{2}{c}{\textbf{\textrm{ETMD}}}  \\
			\cmidrule(r){2-3} \cmidrule(r){4-5}  
			&  CC $\uparrow$ & SIM $\uparrow$ & CC $\uparrow$ & SIM $\uparrow$   \\
			\midrule
			Audio-Only       & 0.343 & 0.283 & 0.365  & 0.295  \\
			Video-Only       & 0.716 & 0.556  & 0.644  & 0.503  \\
			Ours             & \textbf{0.738} & \textbf{0.571} & \textbf{0.652} & \textbf{0.506} \\
			\bottomrule
	\end{tabular}}
	\label{table-ablation_av_encoders}
	\vspace{-10pt}
\end{table}

\begin{figure}[t]
	%	\flushleft
	\centering
	\includegraphics[scale=0.5]{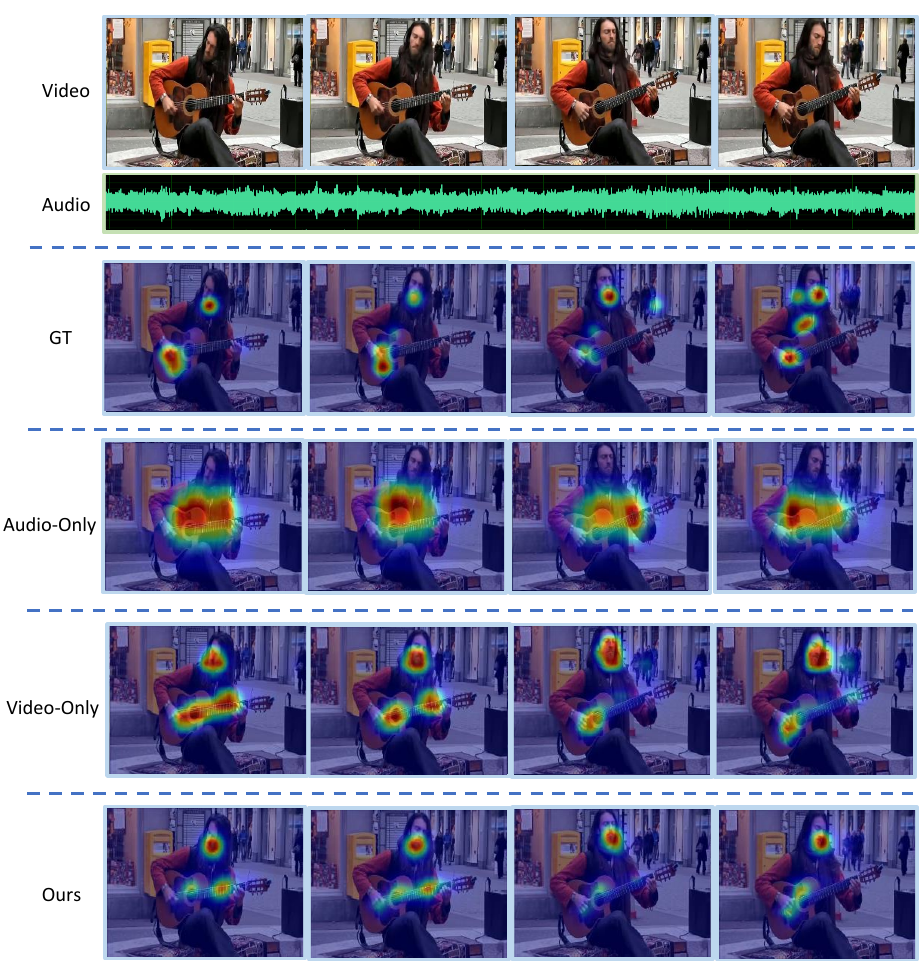}
	\vspace{-10pt}
	\caption{Visualizing the saliency results when different modalities are used. The audio-only approach can localize the sound source coming from the performer's guitar, while the video-only approach focuses on both the performer's face as well as the guitar.}
	\vspace{-10pt}
	\label{fig-ablation-av-encoders} 
\end{figure}

%In the experimental Table \ref{table-ablation_1}, the aim is to investigate the effectiveness of each proposed component. A strong baseline method is first employed which models the video-only version of DiffSal framework, replacing the designed multi-modal attention modulation with the purely convolution operation.

\noindent \textbf{Effect of Components of DiffSal}. 
To validate the effectiveness of each module in the proposed framework, a baseline model is initially defined as the video-only version of DiffSal and replaces the multi-modal attention modulation in Saliency-UNet with a pure convolution operation.
As shown in Table \ref{table-ablation_1}, the baseline model demonstrates a good performance that indicates a potential capability of the diffusion-based framework in the AVSP task. With the incorporation of the designed efficient spatio-temporal cross-attention (ECA), and the multi-modal interaction module (MIM) components, the overall performance of the model has been enhanced continually. As the core module, ECA presents a significant improvement in the CC metric by 0.015 on the AVAD dataset, and 0.012 on the ETMD dataset for the whole DiffSal framework. With the addition of the audio features and the MIM, the model also has another improvement of 0.022 in the CC metric on the AVAD dataset. All of these have demonstrated the effectiveness of ECA and MIM in the proposed DiffSal.

\noindent \textbf{Effect of Video and Audio Modalities}. 
Table \ref{table-ablation_av_encoders} shows the contribution of spatio-temporal information from each modality in the Video and Audio Encoders to the overall performance.
The experimental observations reveal that the video-only model exhibits significantly greater strength than the spatio-temporal audio-only version, which verifies the essential role of the video modality. Figure \ref{fig-ablation-av-encoders} also visualizes the model predictions utilizing the two modal encoders separately. 
It is clear that either the audio-only or video-only approach can predict saliency areas in the scene, and the combination of the two modalities leads to more accurate predictions.
This demonstrates the generalization of the DiffSal framework to audio-only, video-only, and audio-visual scenarios as well.

%This showcases the generalizability  of our DiffSal framework, which proves effective not only in AVSP tasks in multi-modal scenarios but also in uni-modal scenarios where a certain modality is absent.
%Considering that both video and audio features encode spatio-temporal information, we further investigate the contribution of the spatio-temporal information of each modality in the  Video and Audio Encoders to the overall performance in Table \ref{table-ablation_av_encoders}. We observe that the video features alone are much stronger than the spatio-temporal audio features, and that video plays an important role. We also visualize the results for the approach that uses two modal encoders separately. In Figure \ref{fig-ablation-av-encoders}, it can be seen that our method can roughly predict the saliency areas in the scene using only the audio modality, and even sometimes agrees with the predictions of the video-only method. This reflects the fact that our DiffSal framework is not only applicable to AVSP tasks in multi-modal scenarios but also to uni-modal scenarios where a certain modality is missing.

\begin{table}[t]
	\centering
	\renewcommand\arraystretch{1.2}
	\caption{Ablation of different multi-modal interaction methods.}
	\vspace{-5pt}
	\resizebox{0.8 \linewidth}{!}{
		\begin{tabular}{l llll}
			\toprule
			%			\multirow{2}{*}{Method} & \multirow{2}{*}{ECAM} & \multirow{2}{*}{MIM} & \multicolumn{2}{c}{\textbf{AVAD}}  & \multicolumn{2}{c}{\textbf{\textrm{ETMD}}}  \\
			\multirow{2}{*}{Method}  & \multicolumn{2}{c}{\textbf{AVAD}}  & \multicolumn{2}{c}{\textbf{\textrm{ETMD}}} \\
			\cmidrule(r){2-3} \cmidrule(r){4-5}  
			& 		 	CC $\uparrow$    & SIM $\uparrow$ & CC $\uparrow$   & SIM $\uparrow$  \\
			\midrule
			DiffSal(w/ MIM)						& \textbf{0.738} & \textbf{0.571} & \textbf{0.652}  & \textbf{0.506}    \\
			\midrule
			\quad w/ Bilinear					& 0.716			 & 0.556          & 0.644           & 0.503    \\
			\quad w/ Addition					& 0.706			 & 0.543          & 0.606           & 0.464     \\
			\quad w/ Concatenation				& 0.704			 & 0.528          & 0.610           & 0.432   \\
			\bottomrule
	\end{tabular}}
	\label{table-ablation-fusion}
%	\vspace{-5pt}
\end{table}

\begin{table}[t]
	\centering
	\renewcommand\arraystretch{1.2}
	\caption{Ablation of different cross-attention strategies. The computational cost is evaluated based on input audio of size $1 \times 9 \times 112 \times 192$ and video of size  $3 \times 16 \times 224 \times 384$. \# Params and \#Mem denote the number of parameters and memory footprint of the model, respectively.}
		\vspace{-5pt}
	\resizebox{0.9 \linewidth}{!}{
			\begin{tabular}{l llll ll}
					\toprule
					\multirow{2}{*}{Attention}  & \multirow{2}{*}{\#Params} & \multirow{2}{*}{\#Mem} &  \multicolumn{2}{c}{\textbf{AVAD}}  & \multicolumn{2}{c}{\textbf{\textrm{ETMD}}}  \\
					\cmidrule(r){4-5} \cmidrule(r){6-7}  
					& & &  CC $\uparrow$ & SIM $\uparrow$ & CC $\uparrow$ & SIM $\uparrow$   \\
					\midrule
					SCA 	  & 76.43M & 5.32G	& 0.713 & 0.531 & 0.628 & 0.476    \\
					ECA 	  &  76.57M & 1.20G		& \textbf{0.738} & \textbf{0.571} & \textbf{0.652}  & \textbf{0.506}  \\     				
					\bottomrule
					
			\end{tabular}}
		\vspace{-10pt}
	\label{table-ablation_CA}
\end{table}

\noindent \textbf{Effect of Different Cross-Attention Strategies}. The design of efficient spatio-temporal cross-attention mechanism is further evaluated.  As shown in Table \ref{table-ablation_CA},  using efficient spatio-temporal cross-attention (ECA) not only leads to better performance but also greatly reduces the memory footprint of the model compared to using the standard cross-attention (SCA) strategy.
This shows that the designed ECA can compress the effective spatio-temporal cues in the features and reduce the interference of irrelevant noises.

% In practice, video features and noise features are first concatenated, and then these methods are used to interact the processed features with audio features. 
%These multi-modal interaction methods are commonly adopted by recent state-of-the-art approaches \cite{tsiami2020stavis, jain2021vinet}. 

\noindent \textbf{Effect of Different Multi-modal Interaction Methods}. 
The effects of using different multi-modal interaction methods, such as Bilinear \cite{tenenbaum2000separating}, Addition, and Concatenation, are compared in Table \ref{table-ablation-fusion}.
These multi-modal interaction methods can be found in recent state-of-the-art works \cite{tsiami2020stavis, jain2021vinet}, and the video features and noise features are firstly concatenated before their interaction with audio features.
Experimental results show that the proposed MIM can outperform all the other three interaction methods, and obtain more robust multi-modal features. In contrast, the performance degradation of the other three methods suffers from the noise information embedded in the features.

\begin{table}[t]
	\centering
	\renewcommand\arraystretch{1.2}
	\caption{Ablation of different training losses.}
	\vspace{-5pt}
	\resizebox{0.9 \linewidth}{!}{
		\begin{tabular}{l lll llll}
			\toprule
			\multirow{2}{*}{Model}  &  \multicolumn{3}{c}{Losses}  & \multicolumn{2}{c}{\textbf{AVAD}}  & \multicolumn{2}{c}{\textbf{\textrm{ETMD}}}  \\
			\cmidrule(r){2-4} \cmidrule(r){5-6}  \cmidrule(r){7-8}
			& $\mathcal{L}_{CE}$& $\mathcal{L}_{KL}$& $\mathcal{L}_{MSE}$& CC $\uparrow$  & SIM $\uparrow$ & CC $\uparrow$ & SIM $\uparrow$  \\
			\midrule
			
			\multirow{3}{*}{Ours} & $\checkmark$ & & & 0.690 & 0.490 & 0.617  & 0.422     \\
			& & $\checkmark$ & &0.720 & 0.552 & 0.644  & 0.496   \\
			& & & $\checkmark$ & \textbf{0.738} & \textbf{0.571} & \textbf{0.652}  & \textbf{0.506}    \\
			\bottomrule
	\end{tabular}}
	\vspace{-10pt}
	\label{table-ablation_loss}
\end{table}

\begin{figure}[!tbp]
	\hspace{-1cm}
	\flushright
	\begin{minipage}[t]{0.49\linewidth}
		\centering
		\includegraphics[width=\linewidth]{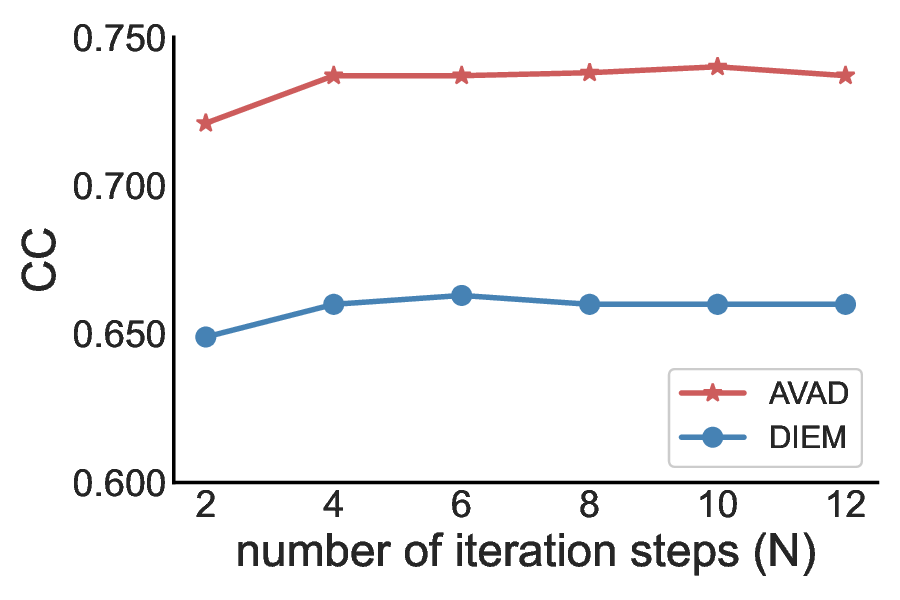}
	\end{minipage}
	\begin{minipage}[t]{0.49\linewidth}
		\centering
		\includegraphics[width=\linewidth]{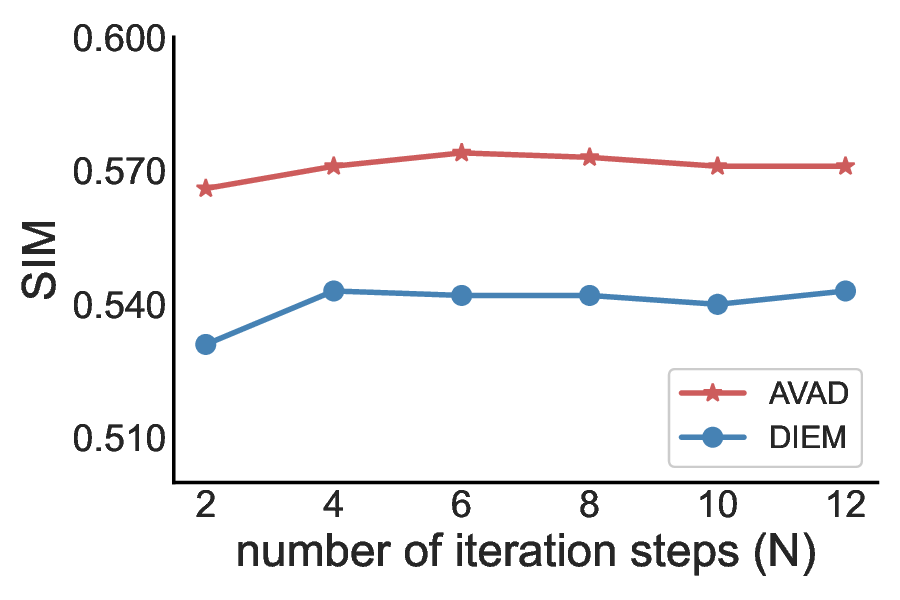}
	\end{minipage}
	\vspace{-5pt}
	\caption{Performance analysis of denoising steps on AVAD and DIEM datasets.}
	\label{fig:exp_denoise_steps}
	\vspace{-10pt}
\end{figure}

\noindent \textbf{Effect of Different Training Losses}. Table \ref{table-ablation_loss} compares the impact  on DiffSal using different loss functions from previous state-of-the-art approaches  \cite{tsiami2020stavis, xiong2023casp}, such as the cross entropy (CE) loss $\mathcal{L}_{CE}$ and the Kullback-Leibler divergence (KL) loss $\mathcal{L}_{KL}$. During the training process,  it is observed that the model with the $\mathcal{L}_{CE}$ converges slowly and yields sub-optimal performance. Compared to the $\mathcal{L}_{MSE}$, employing the  $\mathcal{L}_{KL}$ can achieve acceptable results, but there is still a gap in the performance of training with the $\mathcal{L}_{MSE}$. This suggests that simple MSE loss can be used in the AVSP task as an alternative to these task-tailored loss functions.

\begin{table*}[!htbp]
	\begin{center}
		% Table 1.
		\caption{Comparison with state-of-the-art methods on six audio-visual saliency datasets. \textbf{Bold} text in the table indicates the best result, and \underline{underlined} text indicates the second best result. Our DiffSal significantly outperforms the previous state-of-the-arts by a large margin.}
		\label{table-av-sota}
		\vspace*{-5pt}
		\resizebox{0.95\linewidth}{!}{
			\begin{tabular}{c |c|c | cccc | cccc | cccc}
				\toprule
				\multirow{2}*{\textbf{Method}} & \multirow{2}*{\textbf{\#Params}} & \multirow{2}*{\textbf{\#FLOPs}}&  \multicolumn{4}{c|}{\textbf{DIEM}} & \multicolumn{4}{c|}{\textbf{Coutrot1}} & \multicolumn{4}{c}{\textbf{Coutrot2}}\\
				%				\cmidrule(r){2-5}\cmidrule(r){6-9}\cmidrule(r){10-13}%\cline{3-14}
				\cline{4-15}
				& & & CC $\uparrow$ & NSS $\uparrow$ & AUC-J $\uparrow$  & SIM  $\uparrow$&  CC $\uparrow$ & NSS $\uparrow$ & AUC-J$\uparrow$   & SIM $\uparrow$ & CC$\uparrow$ & NSS$\uparrow$ & AUC-J$\uparrow$  & SIM$\uparrow$ \\
				\hline
				$\textrm{ACLNet}_{\mathit{{\rm TPAMI}^\prime} 2019}$  \cite{wang2019revisiting} & - & - & 0.522 & 2.02 & 0.869 & 0.427 & 0.425 & 1.92 & 0.850  & 0.361 & 0.448 & 3.16 & 0.926 &  0.322 \\
				$\textrm{TASED-Net}_{\mathit{{\rm ICCV}^\prime} 2019}$ \cite{min2019tased} & 21.26M & 91.80G & 0.557 & 2.16 & 0.881 & 0.461 & 0.479 & 2.18 & 0.867  & 0.388 & 0.437 & 3.17 & 0.921 &  0.314 \\
				$\textrm{STAViS}_{\mathit{{\rm CVPR}^\prime} 2020}$ \cite{tsiami2020stavis} & 20.76M & 15.31G & 0.579 & 2.26 & 0.883 & 0.482 & 0.472 & 2.11 & 0.868  & 0.393 & 0.734 & 5.28 & 0.958 & 0.511 \\
				$\textrm{ViNet}_{\mathit{\rm IROS}^{\prime} 2021}$ \cite{jain2021vinet} & 33.97M & 115.31G & 0.632 & 2.53 & 0.899  & 0.498 & 0.56 & 2.73 & 0.889  & 0.425 & 0.754 & 5.95 & 0.951 & 0.493 \\
				$\textrm{TSFP-Net}_{\mathit{\rm arXiv}^{\prime} 2021}$ \cite{chang2021temporal} & - & - & 0.651 & \underline{2.62} & 0.906  & 0.527 & \underline{0.571} & \underline{2.73} & \underline{0.895} & 0.447 & 0.743 & 5.31 & 0.959  & 0.528 \\
				$\textrm{CASP-Net}_{\mathit{{\rm CVPR}^\prime} 2023}$ \cite{xiong2023casp} & 51.62M & 283.35G & \underline{0.655} & 2.61 & \underline{0.906}  & \underline{0.543} & 0.561 & 2.65 & 0.889  & \underline{0.456} & \underline{0.788} & \underline{6.34} & \underline{0.963} & \underline{0.585} \\ 
				%				\midrule
				\hline
				\textbf{Ours}(DiffSal) & 76.57M & 187.31G & \textbf{0.660} & \textbf{2.65} & \textbf{0.907} & \textbf{0.543} & \textbf{0.638} & \textbf{3.20} & \textbf{0.901} & \textbf{0.515} & \textbf{0.835} & \textbf{6.61} & \textbf{0.964} & \textbf{0.625} \\
				%				\bottomrule
				\hline
				\multirow{2}*{\textbf{Method}}  & \multirow{2}*{\textbf{\#Params}} & \multirow{2}*{\textbf{\#FLOPs}} & \multicolumn{4}{c|}{\textbf{AVAD}} & \multicolumn{4}{c|}{\textbf{ETMD}} & \multicolumn{4}{c}{\textbf{SumMe}}\\
				%				\cmidrule(r){2-5}\cmidrule(r){6-9}\cmidrule(r){10-13}
				\cline{4-15}
				& & & CC $\uparrow$ & NSS $\uparrow$ & AUC-J $\uparrow$  & SIM  $\uparrow$&  CC $\uparrow$ & NSS $\uparrow$ & AUC-J$\uparrow$   & SIM $\uparrow$ & CC$\uparrow$ & NSS$\uparrow$ & AUC-J$\uparrow$  & SIM$\uparrow$ \\
				%				\midrule
				\hline
				$\textrm{ACLNet}_{\mathit{{\rm TPAMI}^\prime} 2019}$  \cite{wang2019revisiting} & - & -  & 0.580 & 3.17 & 0.905 & 0.446 & 0.477 & 2.36  & 0.915  & 0.329 & 0.379 & 1.79 & 0.868 &  0.296 \\
				$\textrm{TASED-Net}_{\mathit{{\rm ICCV}^\prime} 2019}$ \cite{min2019tased} & 21.26M & 91.80G& 0.601 & 3.16 & 0.914  & 0.439 & 0.509 & 2.63 & 0.916 & 0.366 & 0.428 & 2.1 & 0.884 & 0.333 \\
				$\textrm{STAViS}_{\mathit{{\rm CVPR}^\prime} 2020}$ \cite{tsiami2020stavis} & 20.76M & 15.31G & 0.608 & 3.18 & 0.919 & 0.457 & 0.569 & 2.94 & 0.931 & 0.425 & 0.422 & 2.04 & 0.888  & 0.337 \\
				$\textrm{ViNet}_{\mathit{\rm IROS}^{\prime} 2020}$ \cite{jain2021vinet} & 33.97M & 115.31G & 0.674 & 3.77 & 0.927 & 0.491 & 0.571 & 3.08 & 0.928 & 0.406 & 0.463 & 2.41 & 0.897  & 0.343 \\
				$\textrm{TSFP-Net}_{\mathit{\rm arXiv}^{\prime} 2021}$  \cite{chang2021temporal} & - & - & \underline{0.704} & 3.77 & 0.932  & 0.521 & 0.576 & 3.07 & 0.932  & 0.428 & 0.464 & 2.30 & 0.894 & 0.360 \\
				$\textrm{CASP-Net}_{\mathit{{\rm CVPR}^\prime} 2023}$ \cite{xiong2023casp} & 51.62M & 283.35G & 0.691 & \underline{3.81} & \underline{0.933} & \underline{0.528} & \underline{0.620} & \underline{3.34} & \underline{0.940} & \underline{0.478} & \underline{0.499} & \underline{2.60} & \underline{0.907} & \underline{0.387} \\ 
				%				\midrule
				\hline
				\textbf{Ours}(DiffSal) & 76.57M & 187.31G & \textbf{0.738} & \textbf{4.22} & \textbf{0.935} & \textbf{0.571} & \textbf{0.652} & \textbf{3.66} & \textbf{0.943} & \textbf{0.506} & \textbf{0.572} & \textbf{3.14} & \textbf{0.921} & \textbf{0.447} \\
				\bottomrule
		\end{tabular}}
	\end{center}
	\vspace{-10pt}
\end{table*}

\begin{figure*}[!tbp]
	\centering
	\vspace*{-5pt}
	\includegraphics[width=0.95 \textwidth]{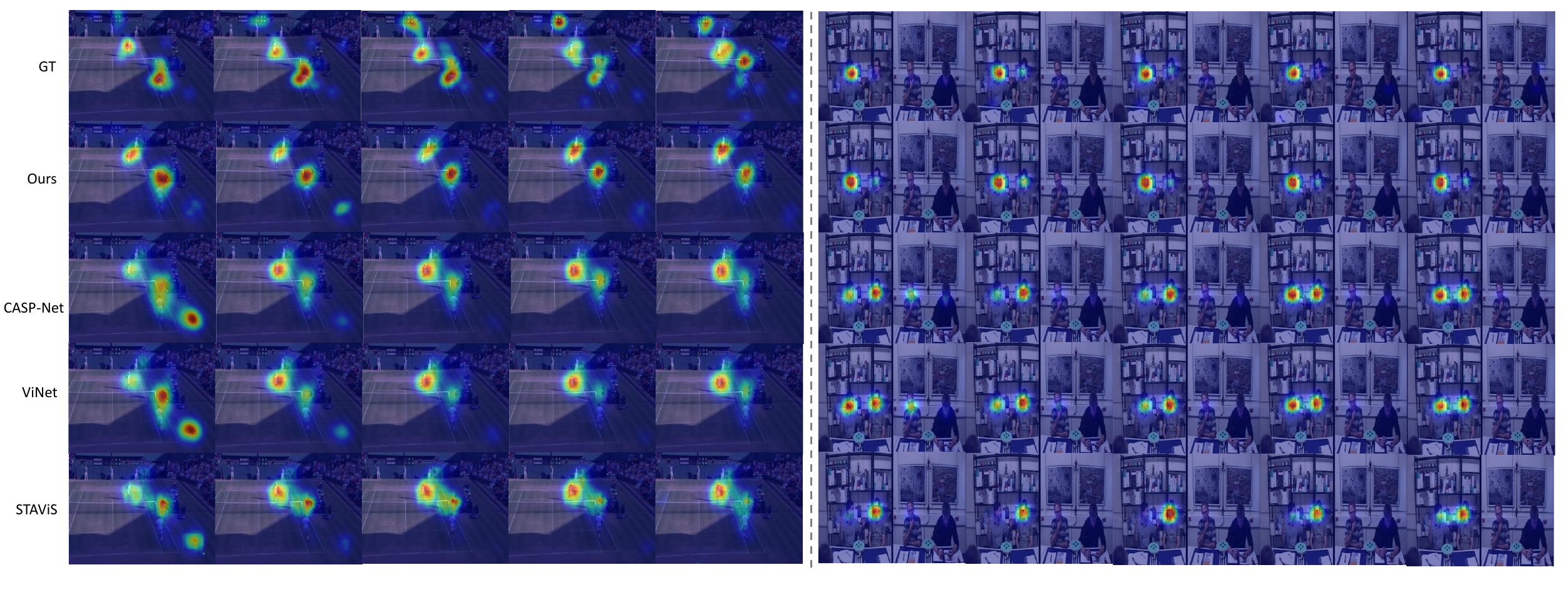}
	\vspace{-5pt}
	\caption{Qualitative results of our method compared with other state-of-the-art works. Challenging scenarios involving fast movement on the tennis court and multiple speakers indoors.
	}
	\label{fig-vis-results}
	\vspace{-10pt}
\end{figure*}

\noindent \textbf{Effect of Denoising Steps}.
The impact of the number of iterative denoising steps on the final performance is studied in Figure \ref{fig:exp_denoise_steps}, which shows that more iteration steps result in better performance. With diminishing marginal benefits as the step number increases, a steady increase in performance is observed. For a linearly increasing of the computational cost with the step number, $N=4$ is used to maintain a good balance between performance and computational cost.

\subsection{Method Comparison}
\noindent \textbf{Comparisons with State-of-the-art Methods}. As shown in Table \ref{table-av-sota}, the experimental results of our DiffSal are compared with recent state-of-the-art works on six audio-visual saliency datasets. The table highlights the superiority of DiffSal, as it outperforms the other comparable works on all datasets by defined metrics. Notably, DiffSal significantly surpasses the previous top-performing methods, such as CASP-Net \cite{xiong2023casp} and ViNet \cite{jain2021vinet}, and becomes the new state-of-the-art on these six benchmarks. The performance boost is very encouraging: DiffSal can achieve an average relative performance improvement of up to 6.3\% compared to the second-place performer. Such substantial improvements validate the effectiveness of the diffusion-based approach as an effective audio-visual saliency prediction framework.

\noindent \textbf{Qualitative Results}. The ability of the model to handle challenging scenarios, such as fast movement on the tennis court and multiple speakers indoors, is further examined.
 Figure \ref{fig-vis-results} compares the DiffSal against other state-of-the-art approaches, such as CASP-Net \cite{xiong2023casp}, ViNet \cite{jain2021vinet} and STAViS \cite{tsiami2020stavis}. It is observed that  DiffSal produces saliency maps much closer to the ground-truth for various challenging scenes. In contrast, CASP-Net focuses mainly on audio-visual consistency and lacks adopting an advanced network structure, leading to sub-optimal results.
STAViS is only able to generate  unsurprisingly saliency maps by employing sound source localization.
More visualization results can be found in the supplementary.

%CASP-Net is only able to roughly localize to salient regions in the scene due to xxx, making it difficult to focus on salient targets in the regions.

\noindent \textbf{Efficiency Analysis}. Table \ref{table-av-sota} compares the number of parameters and computational costs of the DiffSal with previous state-of-the-art works. 
Compared to CASP-Net, the computational complexity of DiffSal is at a moderate level, even though  incorporating Saliency-UNet in DiffSal leads to an increase in the number of model parameters. 
From a performance perspective, the DiffSal model achieves the best performance with the second-highest computational complexity.

%Although the introduction of Saliency-UNet in DiffSal leads to an increase in the number of model parameters compared to CASP-Net, the designed efficient spatio-temporal cross-attention mechanism reduces the computational complexity of DiffSal to a moderate level. 

%Compared to CASP-Net, the introduction of Saliency-UNet in DiffSal leads to an increase in the number of model parameters. However, the designed efficient spatio-temporal cross-attention mechanism reduces the computational complexity of DiffSal to a moderate level. 
%On the one hand, the introduction of Saliency-UNet in DiffSal leads to an increase in the number of model parameters compared to CASP-Net. On the other hand, the design of an efficient cross-attention mechanism reduces the computational complexity of DiffSal to a modest level. From the performance perspective, the DiffSal model achieves optimal performance with the second-highest computational complexity.

\section{Conclusion}

In this work, we introduce a novel Diffusion architecture for generalized audio-visual Saliency prediction (DiffSal), formulating the prediction problem as a conditional generative task of the saliency map by utilizing input video and audio as conditions.
The framework involves extracting spatio-temporal video and audio features from image sequences and corresponding audio signals. A Saliency-UNet is designed to perform multi-modal attention modulation, progressively refining the ground-truth saliency map from the noisy map.
Extensive experiments have proven that DiffSal achieves superior performance compared to previous state-of-the-art methods in six challenging audio-visual benchmarks.

\noindent \textbf{Acknowledgements.} This work was supported by NSFC under Grants 62271239,  61862043, 62171382 and Jiangxi Double Thousand Plan under Grant JXSQ2023201022.

{
    \small
    \bibliographystyle{ieeenat_fullname}
    \bibliography{egbib}
}


\clearpage
\appendix
\setcounter{page}{1}
%\maketitlesupplementary

%\section{Appendix Section}

\section{Training and Inference Algorithms}
\label{sec:rationale}

In this section, we present detailed training and inference algorithms of the proposed DiffSal framework.

\noindent \textbf{Training}.  In the training phase, we perform the diffusion process that corrupts ground-truth saliency maps $S_0$ to noisy maps $S_t$, and train the Saliency-UNet to reverse this process.
Algorithm \ref{alg:training} provides the overall training procedure.

\noindent \textbf{Inference}.  Algorithm \ref{alg:infer} summarizes the detailed inference process of the proposed DiffSal. The parameter $steps$ denotes the number of iterative denoising steps. Specifically, at each sampling step, the Saliency-UNet takes as input random noisy maps or the predicted saliency maps of the last sampling step and outputs the estimated saliency maps of the current step.
We then adopt DDIM to update the heatmaps for the next step.

\begin{algorithm}[!ht]
	\DontPrintSemicolon
	
	\caption{DiffSal Training}
	\label{alg:training}
	
	\KwInput{frames: $I$, audio: $A$, $T$, gt maps: $S_0$}

	  \Repeat{converged}{
	 	 $\textbf{f}_v = \mathrm{\textbf{VideoEncoder}}(I)$;
	 	 
	 	$\textbf{f}_a =\mathrm{\textbf{AudioEncoder}}(A)$;
	 	 	
	 	$t \sim \mathrm{Uniform}({1, ..., T})$;
	 	
	 	$S_t = \sqrt{\bar{\alpha}_t}S_0 + \sqrt{1-\bar{\alpha}_t}\epsilon, \epsilon \in \mathcal{N}(0, \textit{\textbf{I}})$;
	 	
	 	Take gradient descent step on
	 		$\Delta_\theta \| g_\psi(S_t, t, \textbf{f}_a, \textbf{f}_v) - S_0  \|_2^2 $ 	
	 }
\end{algorithm}

\begin{algorithm}[!ht]
	\DontPrintSemicolon
	
	\caption{DiffSal Inference}
	\label{alg:infer}
	
	\KwInput{frames: $I$, audio: $A$, $steps$, $T$}
	\KwOutput{predicted saliency map: $S_{pred}$}

	 $\textbf{f}_v = \mathrm{\textbf{VideoEncoder}}(I)$;

	$\textbf{f}_a =\mathrm{\textbf{AudioEncoder}}(A)$;

	$S_t \sim \mathcal{N}(0, \textit{\textbf{I}})$;

	$\mathrm{times} = \mathrm{Reversed}(\mathrm{Linespace}(-1, T, steps))$;

	$\mathrm{time_{pairs}} = \mathrm{List}(\mathrm{Zip}(\mathrm{times}[:-1], \mathrm{times}[1:]))$;
	
	 \For{$t_{now}, t_{next}$ \KwTo $\mathrm{time_{pairs}}$}{
  		$S_{pred} =  g_\psi(S_t, t_{now}, \textbf{f}_a, \textbf{f}_v)$\;
	  	$S_t = \mathrm{DDIM}(S_t, S_{pred}, t_{now}, t_{next})$
	}
	
\end{algorithm}

\section{Supplementary Experiments}

%In this section, we continue the analysis of DiffSal and evaluate the performance of DiffSal on three video datasets as well as show the visualization results.
This section continues the analysis of DiffSal's components, evaluates DiffSal's performance on three video datasets, and presents visualization results.

\subsection{Further analysis of DiffSal}

\noindent \textbf{Analyzing the Performance of DiffSal using Different Video Encoders}. We conduct experiments within  DiffSal using video encoders employed in other SOTA works, \eg , the 3D ResNet in STAViS and the S3D in CASP-Net, as illustrated in the table below. In comparison to Table \ref{table-encoders}, DiffSal (w/ S3D) surpasses CASP-Net, while DiffSal (w/ 3D ResNet) also outperforms STAViS. This highlights the superiority of our diffusion model-based framework under the same encoders and affirms DiffSal's adaptability to various types of encoders.

\begin{table}[htbp]
%	\vspace{-9pt}
	\centering
	\renewcommand\arraystretch{1.2}
	\resizebox{\linewidth}{!}{
		\begin{tabular}{cl|llllll}
			\toprule
			\multicolumn{2}{c|}{\multirow{2}{*}{\textbf{Method}}} &  \multicolumn{2}{c}{\textbf{AVAD}} & \multicolumn{2}{c}{\textbf{$\mathit{\textrm{ETMD}}$}} & \multicolumn{2}{c}{\textbf{$\mathit{\textrm{Coutrot1}}$}}\\ 
			\cmidrule(r){3-4} \cmidrule(r){5-6}  \cmidrule(r){7-8} 
			\multicolumn{2}{c|}{} &  CC $\uparrow$ & SIM $\uparrow$ & CC $\uparrow$ & SIM $\uparrow$   & CC $\uparrow$ & SIM $\uparrow$  \\ \hline
%			\Xhline{0.5pt}
			\multicolumn{2}{c|}{\textbf{DiffSal}(w/ 3D ResNet)}  & {0.632} & {0.471} & {0.583}  & 0.441  & {0.521}  & 0.417 \\
			
%			\Xhline{0.5pt}
			\multicolumn{2}{c|}{\textbf{DiffSal}(w/ S3D)} & 0.708 &  {0.541} & {0.637} & 0.492  & {0.578}  & 0.469 \\
			\midrule
			\multicolumn{2}{c|}{\textbf{DiffSal}(w/ MViT)} & \textbf{0.738} & \textbf{0.571} & \textbf{0.652}  & \textbf{0.506}  & \textbf{0.638}  & \textbf{0.515} \\
			\bottomrule
	\end{tabular}}
	\caption{Compare the performance of DiffSal using different video encoders.}
	\label{table-encoders}
%	\vspace{-10pt}
\end{table}

\begin{figure}[t]
	\centering
	\includegraphics[scale=0.5]{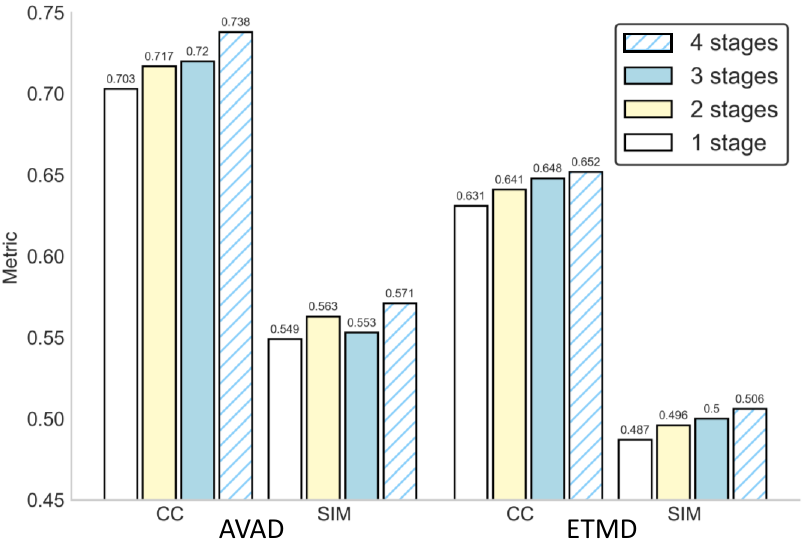}
	\caption{Analyzing the effect of varying the number of multi-modal attention modulation stages  on the AVAD and ETMD datasets.
	}
	\label{fig-transformer_stages}
\end{figure}

\noindent \textbf{Analyzing the Number of Multi-modal Attention Modulation Stages}.
The decoder part of the Saliency-UNet is configured with four stages by default. Figure \ref{fig-transformer_stages} shows the impact of varying the number of multi-modal attention modulation stages on task performance across the AVAD and ETMD datasets. Notably, the most optimal performance is achieved when the number of multi-modal attention modulation stages is set to \textbf{4}. These results imply that Saliency-UNet benefits from progressively fusing audio and video features at multiple scales.

\begin{figure*}[!tbp]
	\centering
	
	\includegraphics[width= \textwidth]{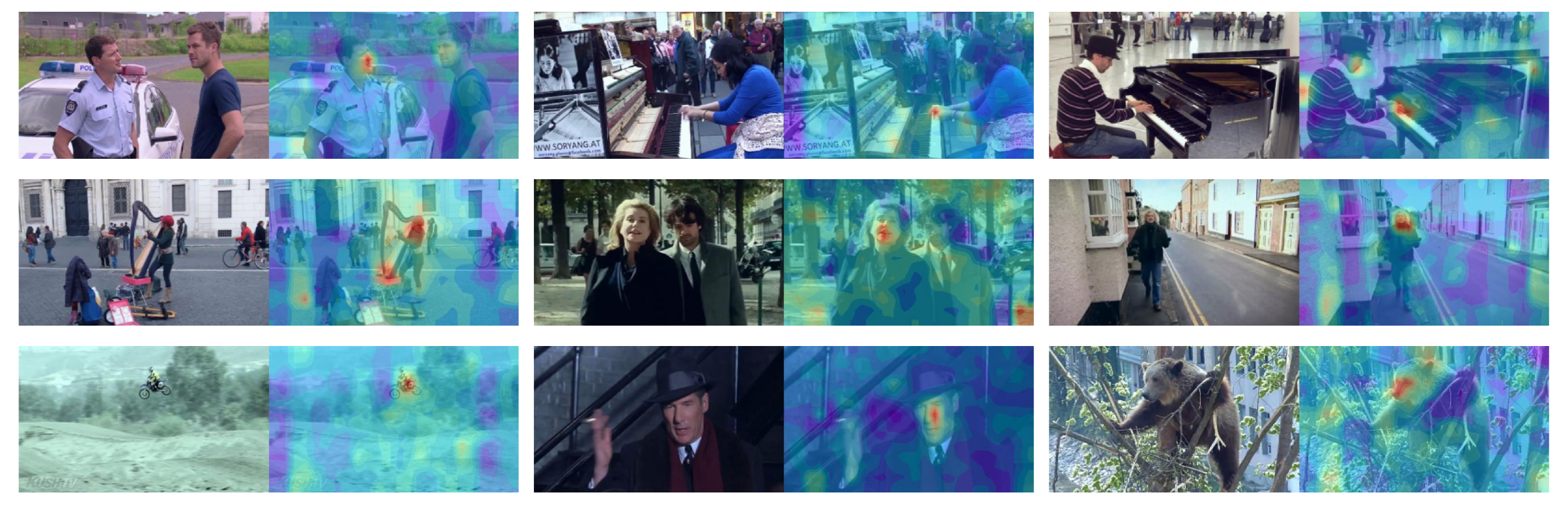}
	\caption{Visualizing the key audio-visual activity features learned by multi-modal interaction module when generating saliency maps. Each pair of pictures shows the frame of the sounding object in the scene (left) and the key audio-visual activity area overlaid (right).
	}
	\label{fig-av-appendix}
	
\end{figure*}

\noindent \textbf{Visualizing Key Audio-Visual Activities}. Figure \ref{fig-av-appendix} illustrates the key audio-visual activity features learned by the multi-modal interaction module during the generation of the saliency maps in DiffSal.  It is obvious that the highlighted  key audio-visual activity regions correspond well to the sound sources in the frame.  For example, the DiffSal model can focus on the main speaker in two-person dialog scenes with the help of sound, and attend to the position of musical instruments in playing scenes. 
This further confirms the ability of the proposed  multi-modal interaction module to capture key audio-visual activity regions, which in turn enhances saliency prediction performance.

\subsection{Comparison with Video Saliency Prediction Methods}

For comprehensive validation, the performance of the video-only version of the DiffSal model is analyzed on three commonly used video datasets: DHF1k \cite{wang2018revisiting}, Hollywood2 \cite{marszalek09}, and UCF-Sports \cite{peng2018two}.
(i) DHF1k comprises 600 training videos, 100 validation videos, and 300 testing videos, all with a frame rate of 30 fps. The DiffSal model can only be evaluated on the validation set of DHF1k due to unavailable annotations of the test set, following \cite{ma2022video, hu2023tinyhd}.
(ii) Hollywood2 consists of 1707 videos extracted from 69 movies, with 12 categorized action classes. For training, 823 videos are used, and for testing, 884 videos are utilized.
(iii) UCF-Sports contains 150 videos, with 103 for training and 47 for testing. These videos are collected from broadcast TV channels and cover 9 sports, including diving, weightlifting, and horse riding.

\begin{table*}[h]
	\begin{center}
		% Table 2.
		\centering
		\renewcommand\arraystretch{1.2}
		\caption{Comparison with state-of-the-art methods on three video datasets. \textbf{Bold} text in the table indicates the best result, and \underline{underlined} text indicates the second best result. Our DiffSal is comparable to the previous state-of-the-arts.}
		\label{table-video-sota}
		\resizebox{0.95\linewidth}{!}{
			\begin{tabular}{c| c| c| cccc| cccc| cccc}
				\toprule
				\multirow{2}*{\textbf{Method}} & \multirow{2}*{\textbf{\#Params}} & \multirow{2}*{\textbf{\#FLOPs}} & \multicolumn{4}{c|}{\textbf{DHF1k}} & \multicolumn{4}{c|}{\textbf{Hollywood2}} & \multicolumn{4}{c}{\textbf{UCF-Sports}}\\
				\cline{4-15}
				& & & CC $\uparrow$ & NSS $\uparrow$ & AUC-J $\uparrow$  & SIM  $\uparrow$&  CC $\uparrow$ & NSS $\uparrow$ & AUC-J$\uparrow$   & SIM $\uparrow$ & CC$\uparrow$ & NSS$\uparrow$ & AUC-J$\uparrow$  & SIM$\uparrow$ \\				
				\hline
				$\textrm{TASED-Net}_{\mathit{\rm ICCV}^{\prime} 2019}$\cite{min2019tased} 			& 21.26M & 91.80G & 0.440 & 2.541 & 0.898 & 0.351   & 0.646 & 3.302 & 0.918  & 0.507 & 0.582 & 2.920 & 0.899 & 0.469 \\
				$\textrm{UNIVSAL}_{\mathit{\rm ECCV}^{\prime} 2020}$\cite{droste2020unified} 		& 3.66M & 14.82G
				& 0.431 & 2.435 & 0.900  & 0.344  & 0.673 & 3.901 & 0.934  & 0.542 & 0.644 & 3.381 & 0.918  & 0.523 \\
				$\textrm{ViNet}_{\mathit{\rm IROS}^{\prime} 2020}$\cite{jain2021vinet} 				& 31.10M & 115.26G & 0.460 & 2.557 & 0.900  & 0.352  & 0.693 & 3.730 & 0.930  & 0.550 & 0.673 & 3.620 & 0.924  & 0.522 \\
				$\textrm{VSFT}_{\mathit{\rm TCSVT}^{\prime} 2021}$\cite{ma2022video} 				& 14.11M & 60.16G & 0.462 & 2.583 & 0.901  & 0.360  & 0.703 & 3.916 & 0.936  & 0.577 & - & - & -  & -   \\
				$\textrm{ECANet}_{\mathit{\rm NeuroComputing}^{\prime} 2022}$\cite{xue2022ecanet} 	& - & -  & - & - & -  & - & 0.673 & 3.380 & 0.929  & 0.526 & 0.636 & 3.189 &  0.917  & 0.498 \\
				$\textrm{STSANet}_{\mathit{\rm TMM}^{\prime} 2022}$\cite{wang2021spatio}			 & -& - & - & - & -  & -  & 0.721 & 3.927 & 0.938  & 0.579 & \underline{0.705} & \textbf{3.908} &  \underline{0.936}  & \underline{0.560} \\
				$\textrm{TinyHD-S}_{\mathit{\rm WACV}^{\prime} 2023}$\cite{hu2023tinyhd} 			& 3.92M & 40.22G  &  0.492 & 2.873 & 0.907 & 0.388  & 0.690 & 3.815 & 0.935  & 0.561 & 0.624 & 3.280 & 0.918 & 0.510 \\
				$\textrm{TMFI-Net}_{\mathit{\rm TCSVT}^{\prime} 2023}$\cite{zhou2023transformer}  	& 53.41M & 305.15G & \underline{0.524} & \underline{3.006} & \underline{0.918} & \textbf{0.410} & \underline{0.739} & \textbf{4.095} & \underline{0.940}  & \underline{0.607} & \textbf{0.707} & \underline{3.863} &  \textbf{0.936}  & \textbf{0.565} \\
				\hline
				\textbf{ Our}(DiffSal)  & 70.54M & 161.06G & \textbf{0.533} & \textbf{3.066} & \textbf{0.918} & \underline{0.405} & \textbf{0.765} & \underline{3.955} & \textbf{0.951} & \textbf{0.610} & 0.685 & 3.483 & 0.928  & 0.543 \\
				\bottomrule
		\end{tabular}}
	\end{center}
\end{table*}

Table \ref{table-video-sota} shows a comparison of the video-only version of DiffSal method against existing state-of-the-arts, including TMFI-Net \cite{zhou2023transformer}, TinyHD-S \cite{hu2023tinyhd}, and STSANet \cite{wang2021spatio}, on three video datasets. Our approach advances the most state-of-the-art methods by an evident margin on DHF1k and Hollywood2 and achieves good performance on UCF-Sports. 
Compared to TMFI-Net, the CC performance of DiffSal improves from 0.524 to 0.533 on DHF1k and from 0.739 to 0.765 on Hollywood2, respectively.
As for the number of parameters and computational complexity of the model, DiffSal has the highest number of parameters, but only about half the computational complexity of the second place TMFI-Net.

The size of the UCF-Sports dataset is minimal compared to the DHF1k and Hollywood datasets, with only 150 videos. Training on the UCF-Sports  dataset causes DiffSal with more parameters to be difficult to converge  completely, and only achieves a sub-optimal state. While other models with less number of parameters are easier to fully optimize on the UCF-Sports dataset.
These experimental results show that the DiffSal model achieves a balance between performance and computational complexity.

\subsection{More Qualitative Analysis}

\begin{figure*}[!tbp]
	\centering
	
	\includegraphics[width= \textwidth]{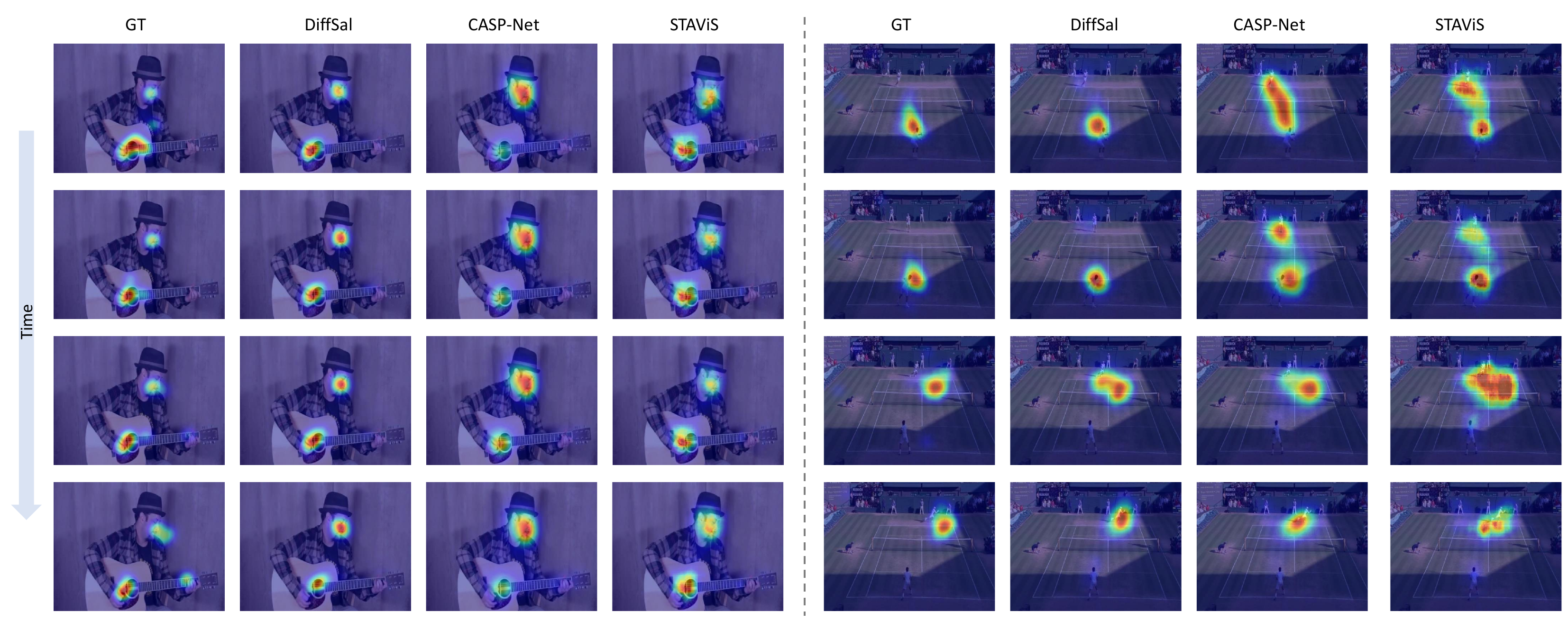}
	\caption{Comparison of visualized saliency maps from the ground-truth, our DiffSal, and previous state-of-the-art CASP-Net and STAViS.
	}
	\label{fig-vis-appendix-1}
	
\end{figure*}

\begin{figure*}[!tbp]
	\centering
	
	\includegraphics[width= \textwidth]{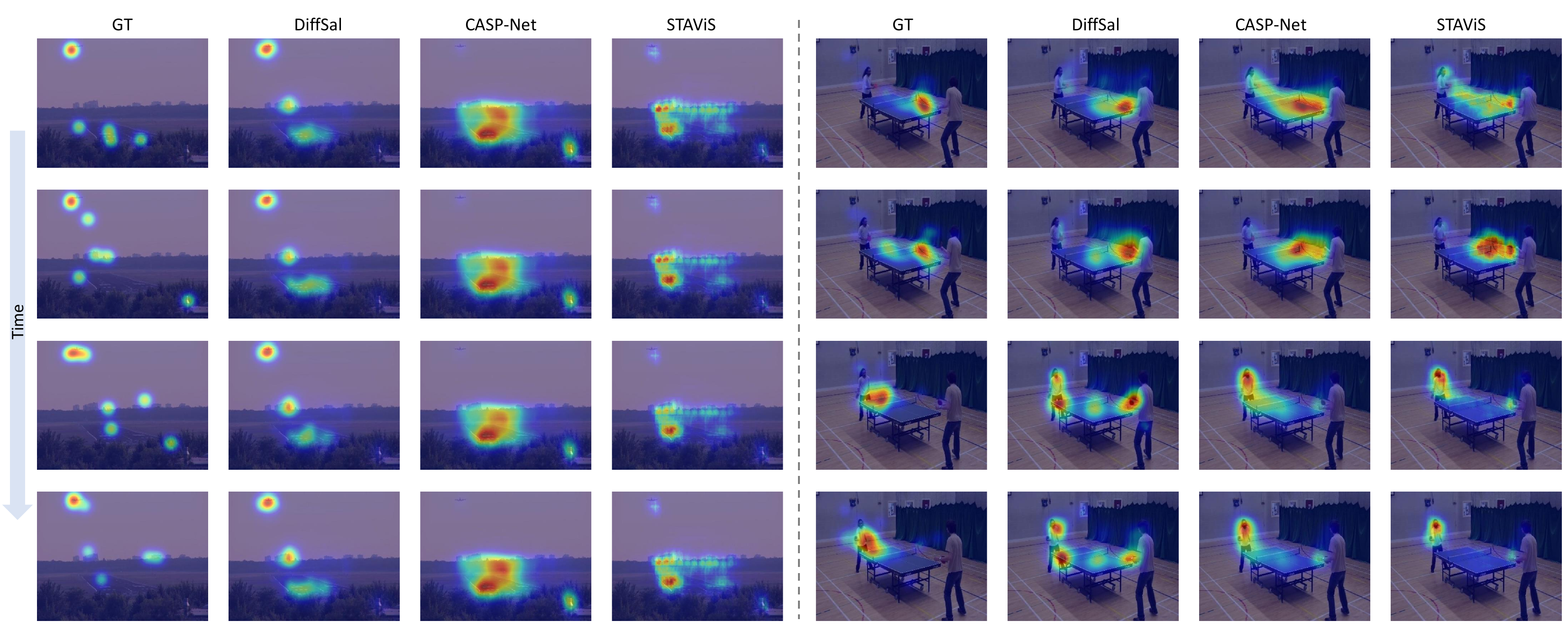}
	\caption{Comparison of visualized saliency maps from the ground-truth, our DiffSal, and previous state-of-the-art CASP-Net and STAViS.
	}
	\label{fig-vis-appendix-2}
	
\end{figure*}

Figures \ref{fig-vis-appendix-1} and \ref{fig-vis-appendix-2} show the performance of DiffSal in diverse real-world scenarios, respectively. These visualizations demonstrate that DiffSal's predictions are much closer to the ground-truth maps, whereas the CASP-Net and STAViS methods struggle to  predict the accurate saliency regions.

\section{Limitations and Future Work}

While DiffSal provides an effective and generalized diffusion-based approach for audio-visual saliency prediction, it also increases the number of parameters and computational complexity of the model.
Exploring ways to lighten the model can further enhance its applicability, \eg, to edge devices with limited computational power.

%\begin{itemize}
%\item The supplementary can back-reference sections of the main paper, for example, we can refer to \cref{sec:intro};
%\item The main paper can forward reference sub-sections within the supplementary explicitly (e.g. referring to a particular experiment); 
%\item When submitted to arXiv, the supplementary will already included at the end of the paper.
%\end{itemize}
%% 
%To split the supplementary pages from the main paper, you can use \href{https://support.apple.com/en-ca/guide/preview/prvw11793/mac#:~:text=Delete%20a%20page%20from%20a,or%20choose%20Edit%20%3E%20Delete).}{Preview (on macOS)}, \href{https://www.adobe.com/acrobat/how-to/delete-pages-from-pdf.html#:~:text=Choose%20%E2%80%9CTools%E2%80%9D%20%3E%20%E2%80%9COrganize,or%20pages%20from%20the%20file.}{Adobe Acrobat} (on all OSs), as well as \href{https://superuser.com/questions/517986/is-it-possible-to-delete-some-pages-of-a-pdf-document}{command line tools}.

\end{document}